\documentclass[10pt,twocolumn,letterpaper,dvipsnames]{article}

\usepackage{iccv}
\usepackage{times}
\usepackage{epsfig}
\usepackage{graphicx}
\usepackage{amsmath}
\usepackage{amssymb}
\usepackage{colortbl}
\usepackage{booktabs} %
\usepackage{multirow}
\usepackage{makecell}
\usepackage{float}
\usepackage{enumitem} %
\usepackage{arydshln}
\usepackage[table,dvipsnames]{xcolor}
\usepackage{bbding}
\usepackage{animate}
\usepackage{grffile}
\usepackage{siunitx} %
\usepackage{tabularx}
\usepackage{bbm}
\usepackage{caption} 
\usepackage{subcaption}
\usepackage{pifont}     %

\usepackage[breaklinks=true,bookmarks=false]{hyperref}

\iccvfinalcopy %

\def\iccvPaperID{****} %

\ificcvfinal\fi

\newcommand{\bc}{\mathbf{c}}

\newcommand{\bn}{\mathbf{n}}
\newcommand{\bo}{\mathbf{o}}

\newcommand{\br}{\mathbf{r}}

\newcommand{\bv}{\mathbf{v}}

\newcommand{\bx}{\mathbf{x}}

\newcommand{\bz}{\mathbf{z}}

\newcommand{\nR}{\mathbb{R}}

\newcommand{\cK}{\mathcal{K}}
\newcommand{\cL}{\mathcal{L}}

\newcommand{\cR}{\mathcal{R}}

\newcommand{\cX}{\mathcal{X}}

\newcommand{\figref}[1]{Fig.~\ref{#1}}
\newcommand{\secref}[1]{Sec.~\ref{#1}}

\newcommand{\eqnref}[1]{Eq.~\eqref{#1}}
\newcommand{\tabref}[1]{Table~\ref{#1}}

\makeatletter
\DeclareRobustCommand\onedot{\futurelet\@let@token\@onedot}
\def\@onedot{\ifx\@let@token.\else.\null\fi\xspace}

\def\cf{cf\onedot}

\def\wrt{wrt\onedot}

\makeatother

\newcommand{\boldparagraph}[1]{\vspace{0.5em}\noindent{\bf #1.}}
\newcommand{\boldparagraphnoperiod}[1]{\vspace{0.5em}\noindent{\bf #1}}
\renewcommand{\paragraph}[1]{\boldparagraph{#1}}
\newcommand{\paragraphcolon}[1]{\boldparagraphnoperiod{#1:}}

\definecolor{darkgreen}{rgb}{0,0.7,0}
\definecolor{newyellow}{rgb}{1,0.8,0.05}
\definecolor{newgreen}{rgb}{0.2,0.8,0.2}

\makeatletter
\def\adl@drawiv#1#2#3{%
        \hskip.5\tabcolsep
        \xleaders#3{#2.5\@tempdimb #1{1}#2.5\@tempdimb}%
                #2\z@ plus1fil minus1fil\relax
        \hskip.5\tabcolsep}
\newcommand{\cdashlinelr}[1]{%
  \noalign{\vskip\aboverulesep
           \global\let\@dashdrawstore\adl@draw
           \global\let\adl@draw\adl@drawiv}
  \cdashline{#1}
  \noalign{\global\let\adl@draw\@dashdrawstore
           \vskip\belowrulesep}}
\makeatother

\newcommand{\figpipeline}{
\begin{figure*}[ht]
  \centering
  \small
  \includegraphics[width=\linewidth]{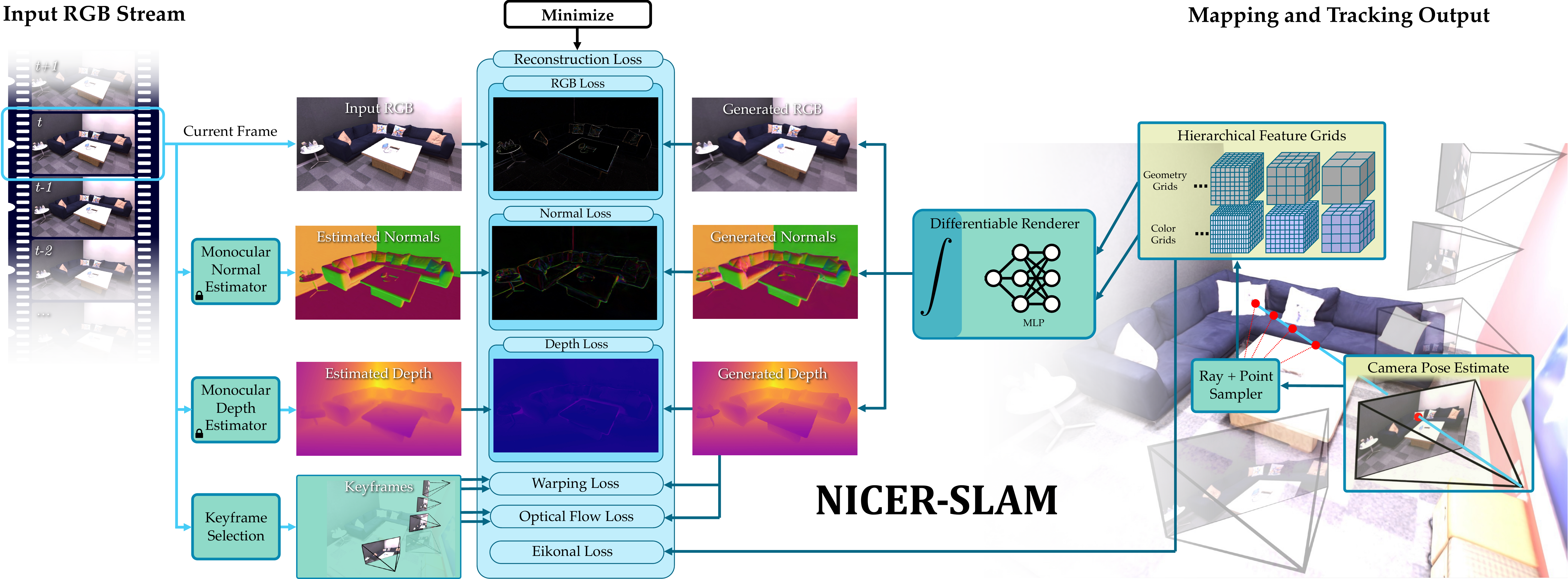}\\
  [+1pt]
  \caption{\textbf{System Overview.} Our method takes only an RGB stream as input and outputs both the camera poses as well as a learned hierarchical scene representation for geometry and colors.
  To realize an end-to-end joint mapping and tracking, we render predicted colors, depths, normals and optimize \wrt the input RGB and monocular cues. Moreover, we further enforce the geometric consistency with an RGB warping loss and an optical flow loss.
  }
  \label{fig:system_overview}
\end{figure*}
}

\newcommand{\figreplicareconstruction}{
\begin{figure*}[!t]
 \vspace{-25pt}
  \centering
  \footnotesize
  \setlength{\tabcolsep}{1.5pt}
  \newcommand{\sz}{0.162}
  \begin{tabular}{cc|ccc|c}
    \includegraphics[width=\sz\linewidth]{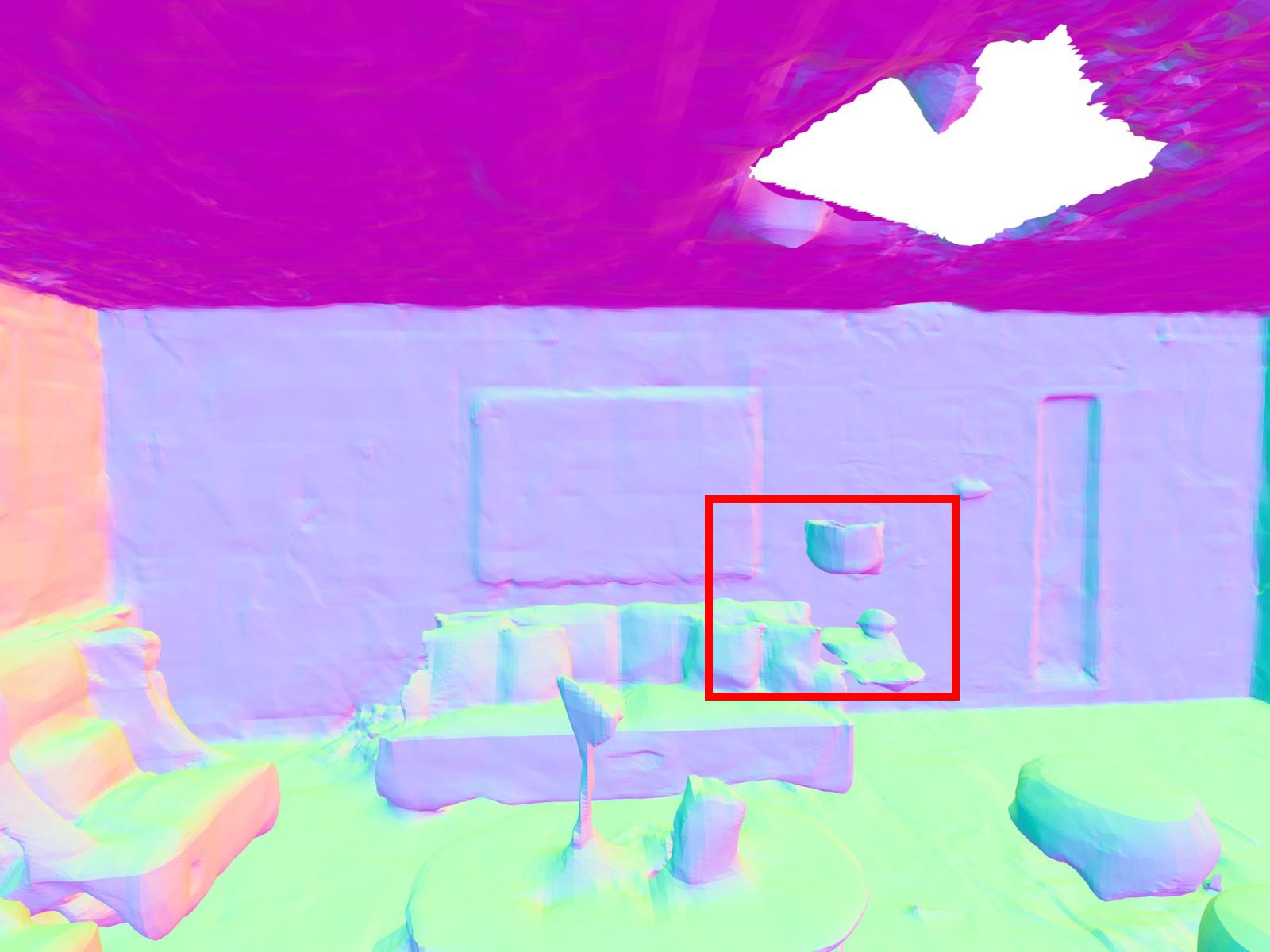} &
    \includegraphics[width=\sz\linewidth]{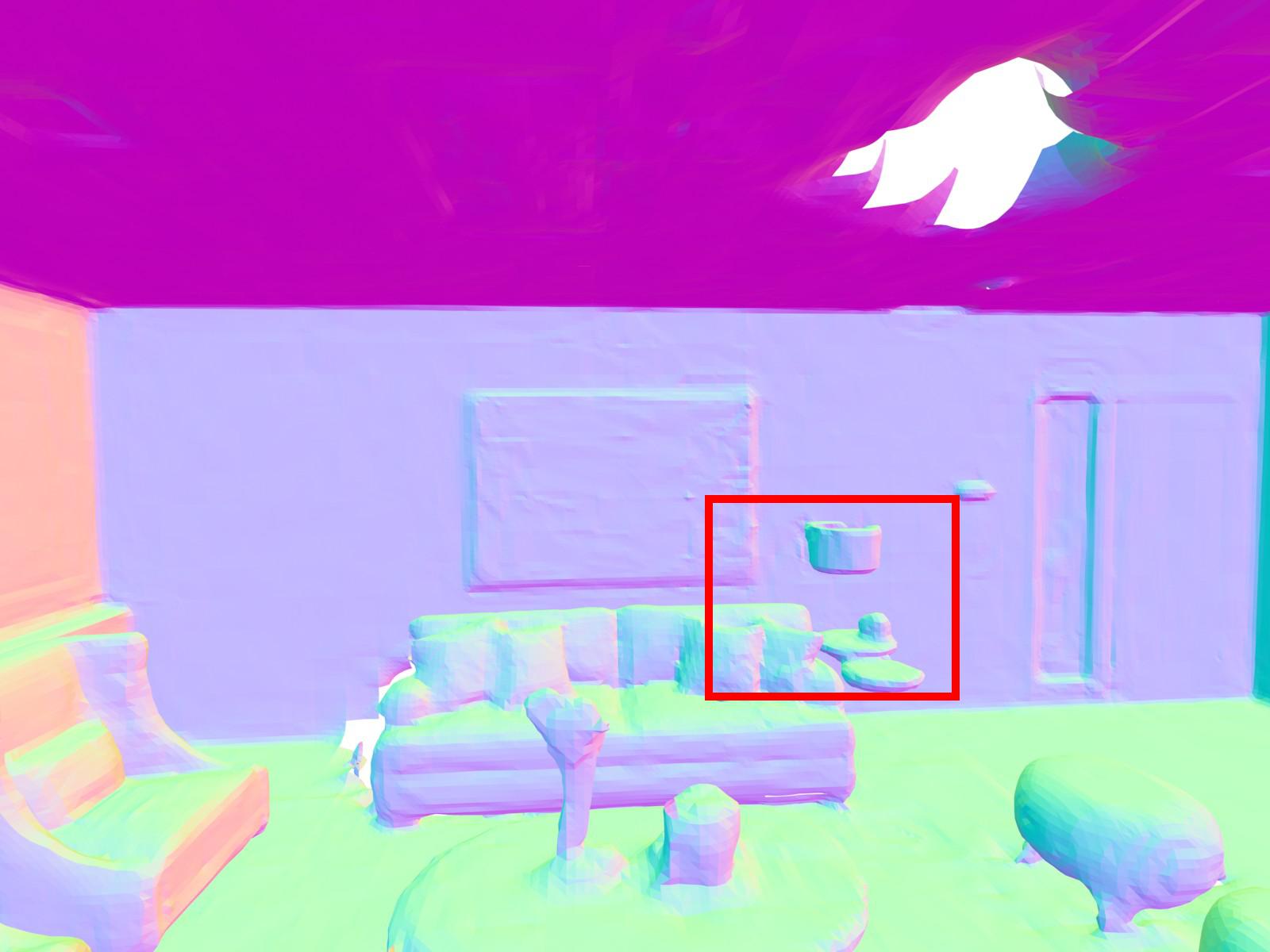} & \includegraphics[width=\sz\linewidth]{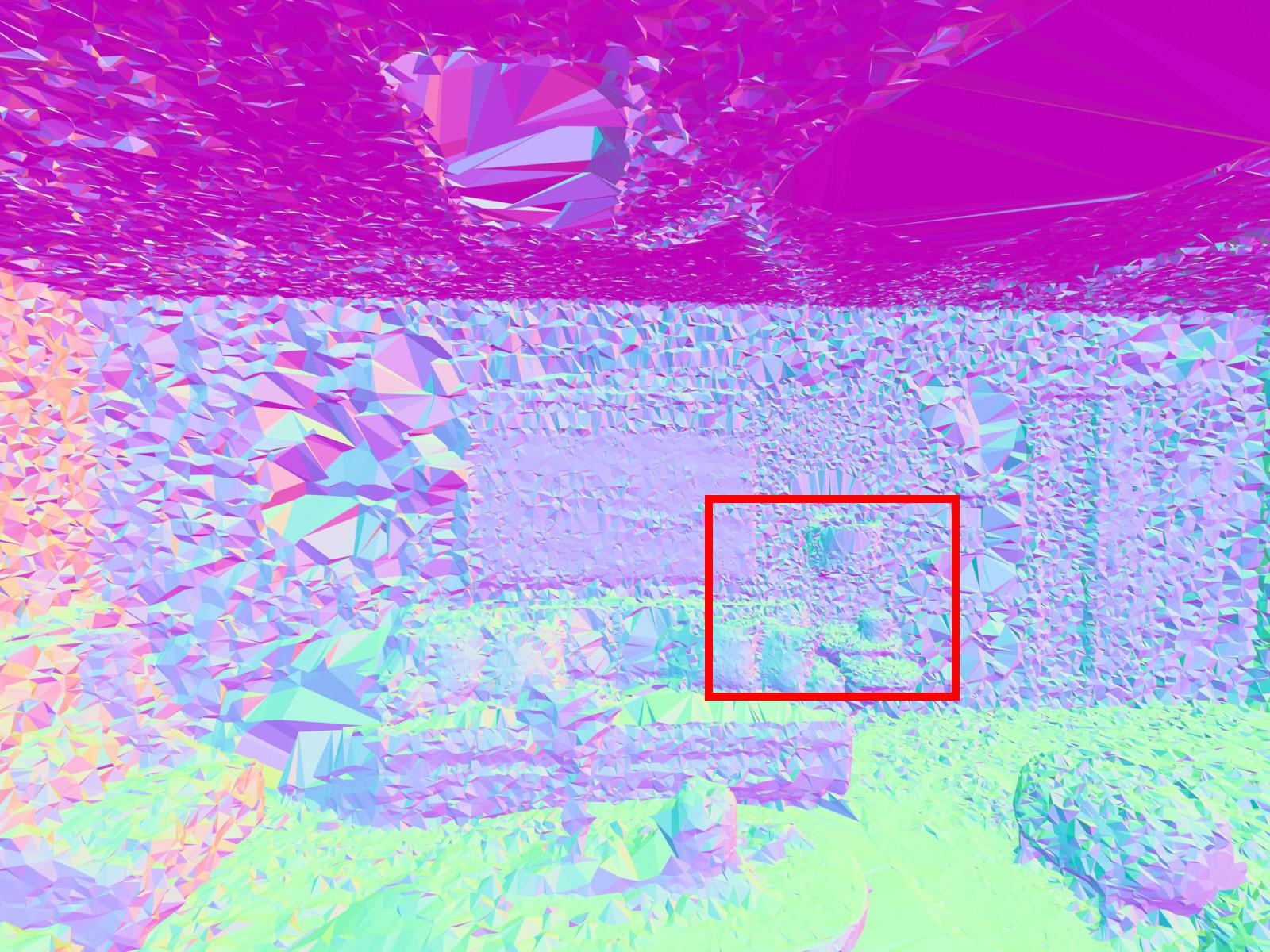} &
    \includegraphics[width=\sz\linewidth]{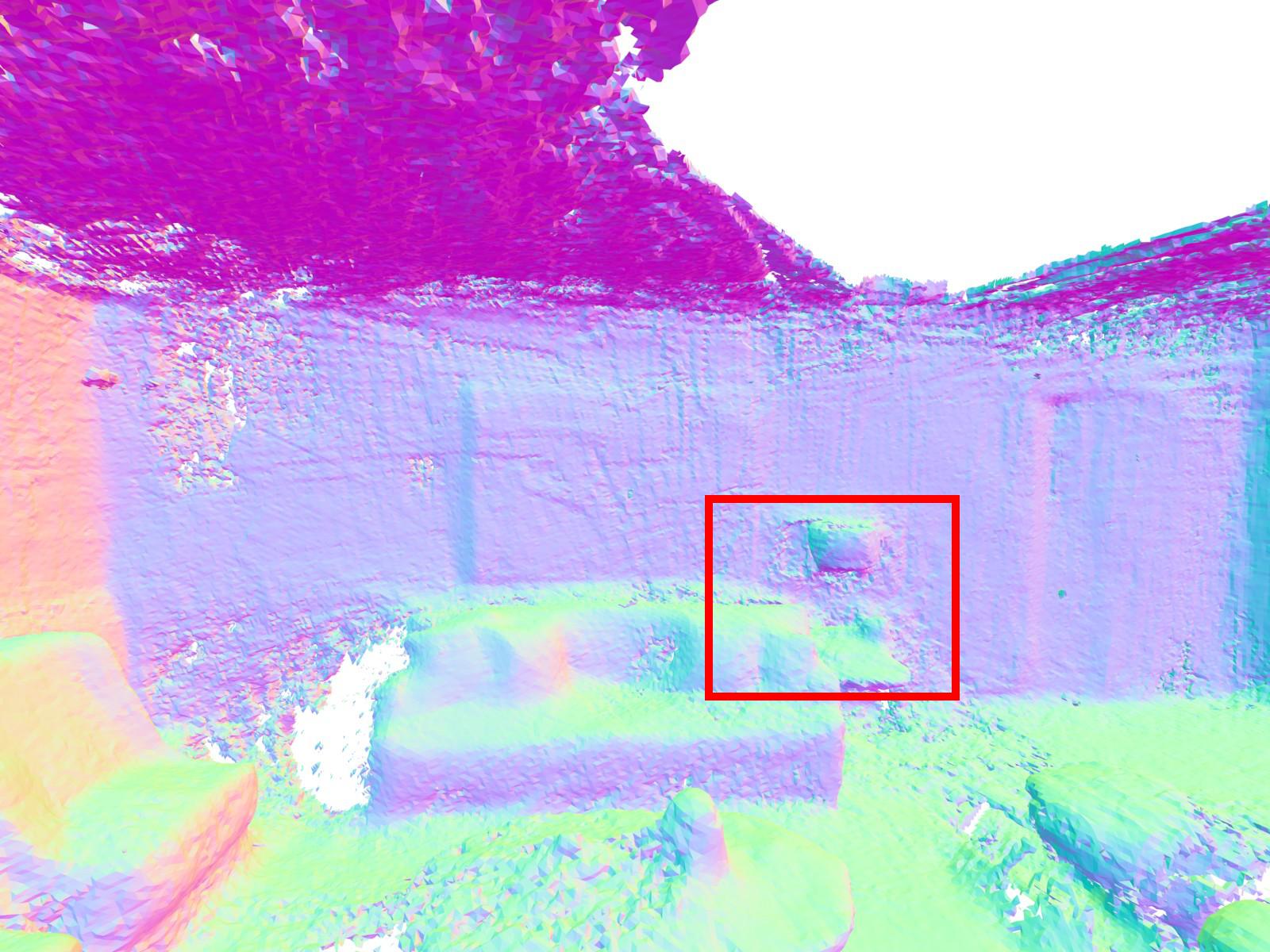} &
    \includegraphics[width=\sz\linewidth]{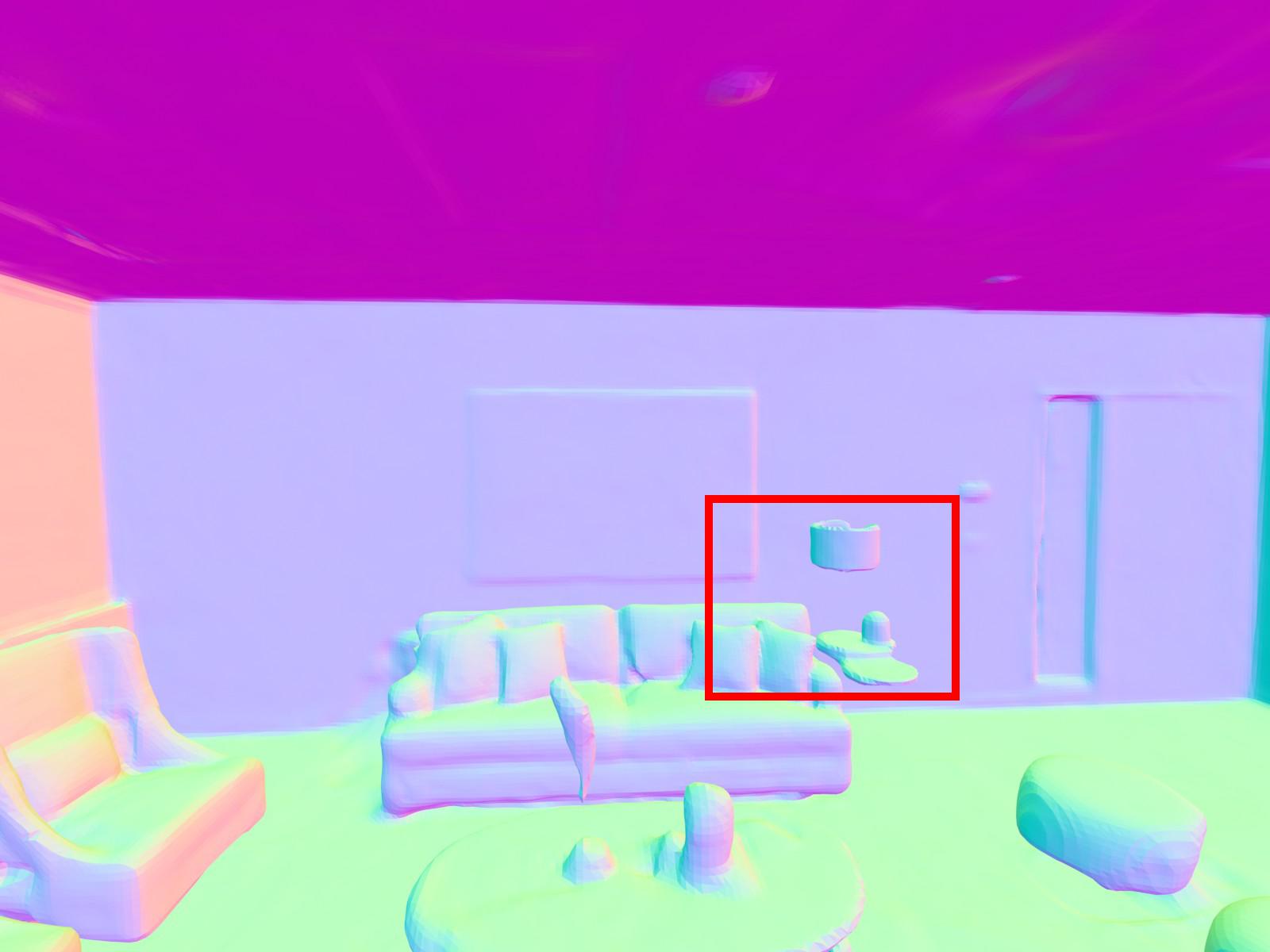} &
    \includegraphics[width=\sz\linewidth]{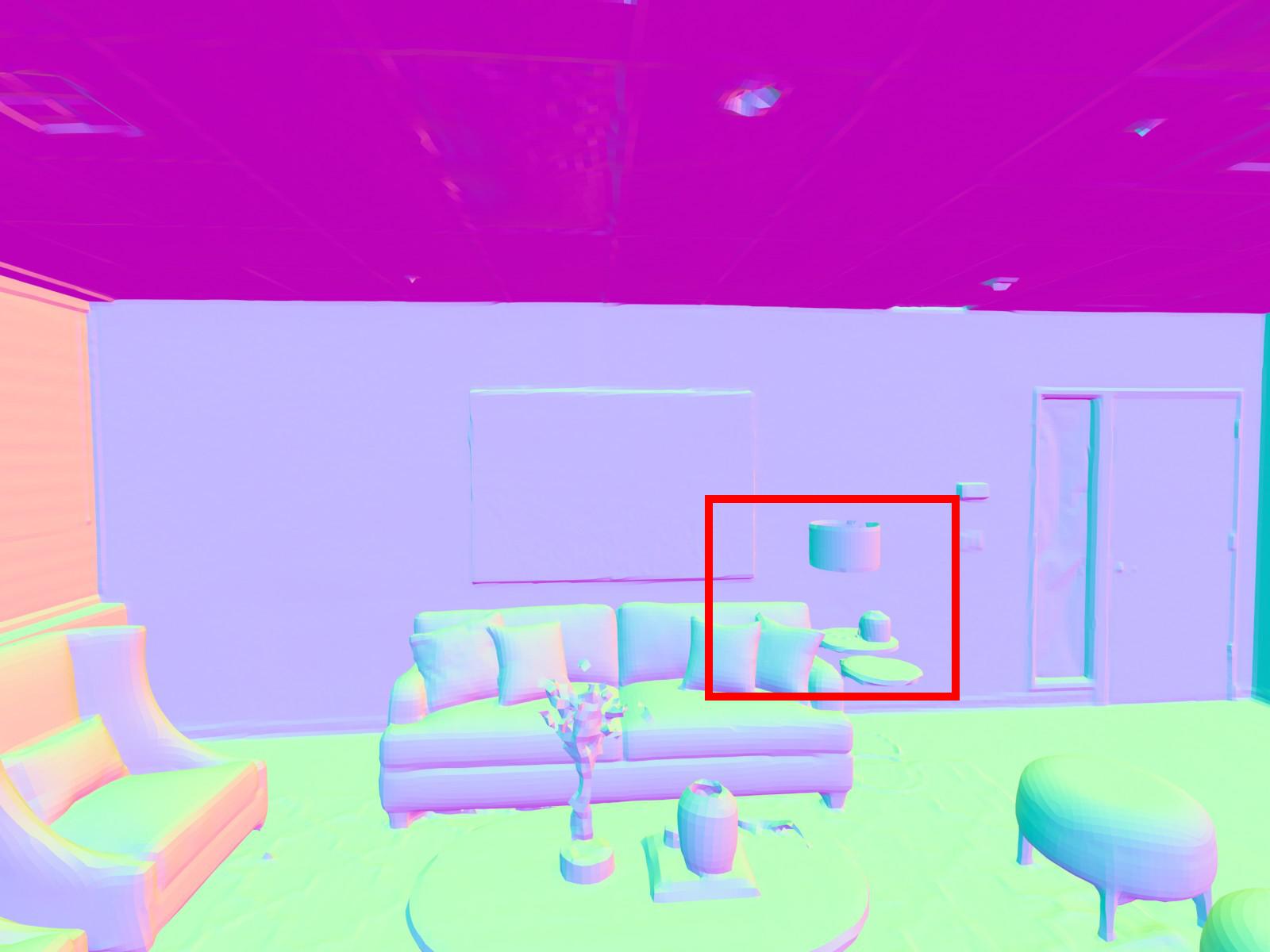} \\
    \includegraphics[width=\sz\linewidth]{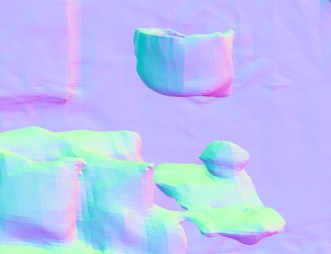} &
    \includegraphics[width=\sz\linewidth]{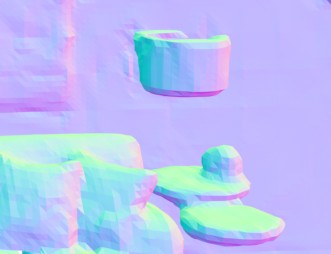} & 
    \includegraphics[width=\sz\linewidth]{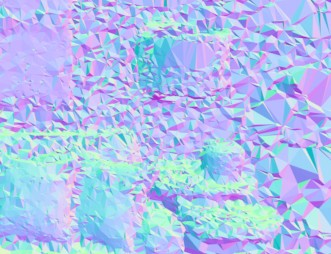} & \includegraphics[width=\sz\linewidth]{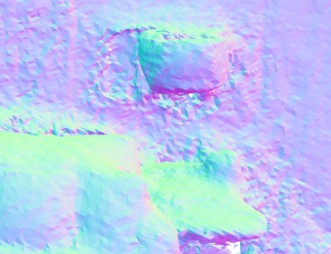} &
    \includegraphics[width=\sz\linewidth]{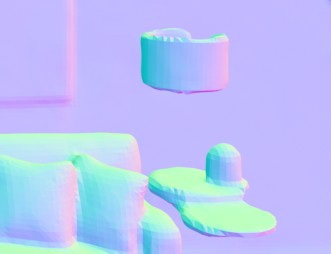} &
    \includegraphics[width=\sz\linewidth]{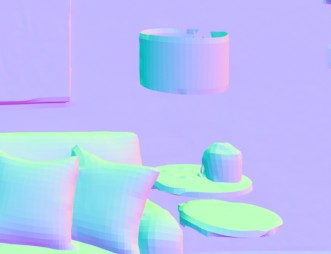} \\
    \includegraphics[width=\sz\linewidth]{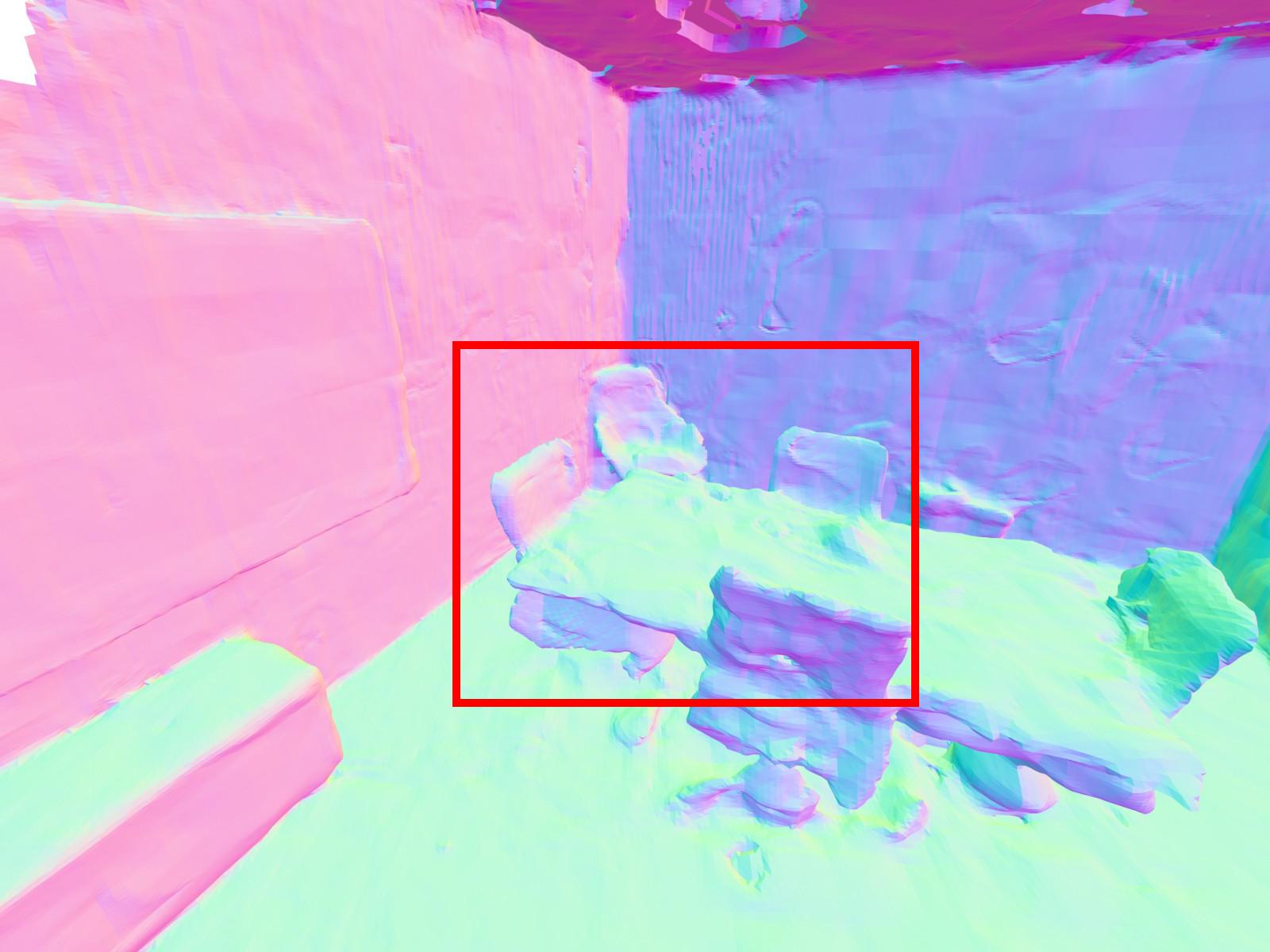} &
    \includegraphics[width=\sz\linewidth]{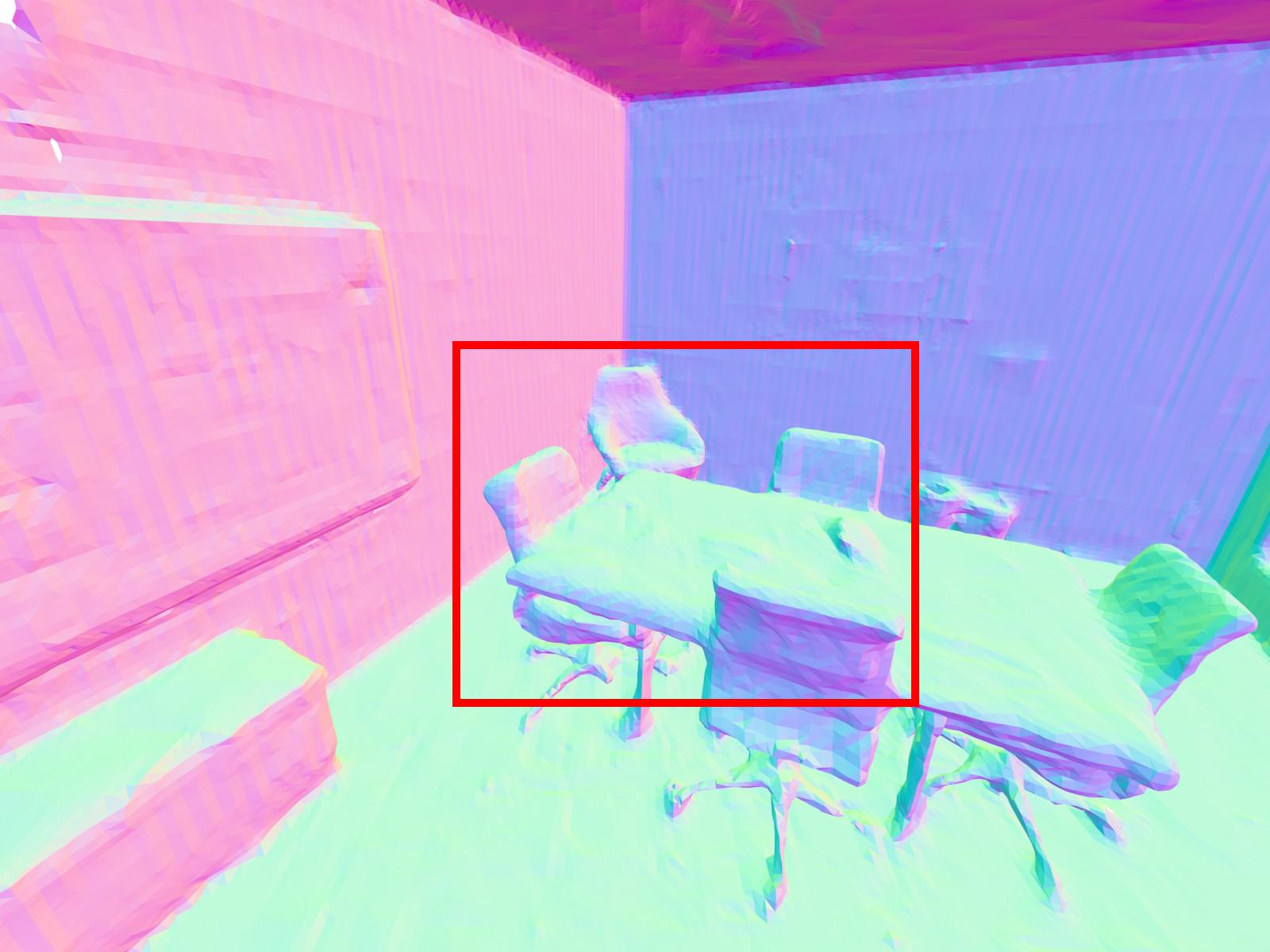} & 
    \includegraphics[width=\sz\linewidth]{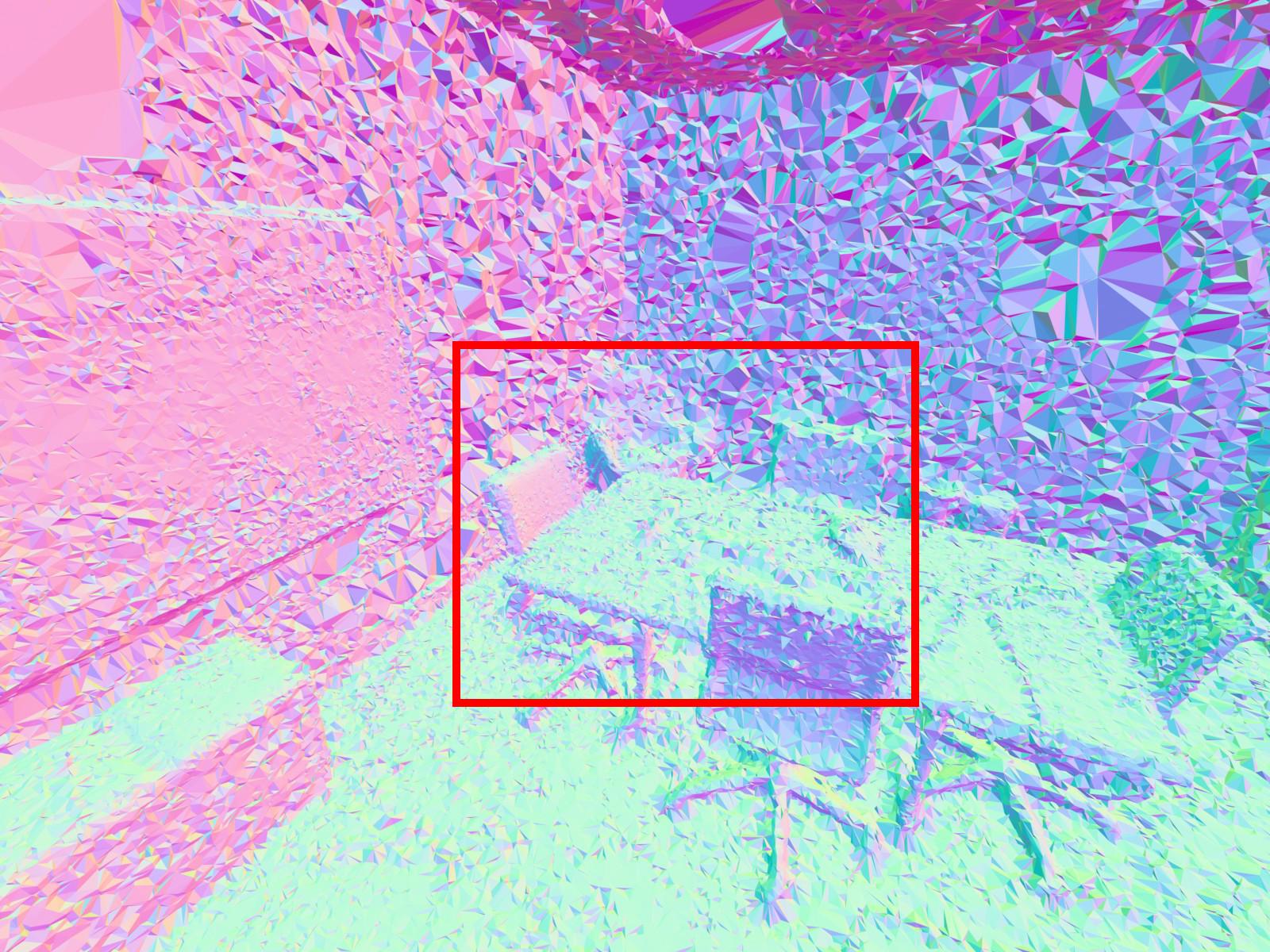} & \includegraphics[width=\sz\linewidth]{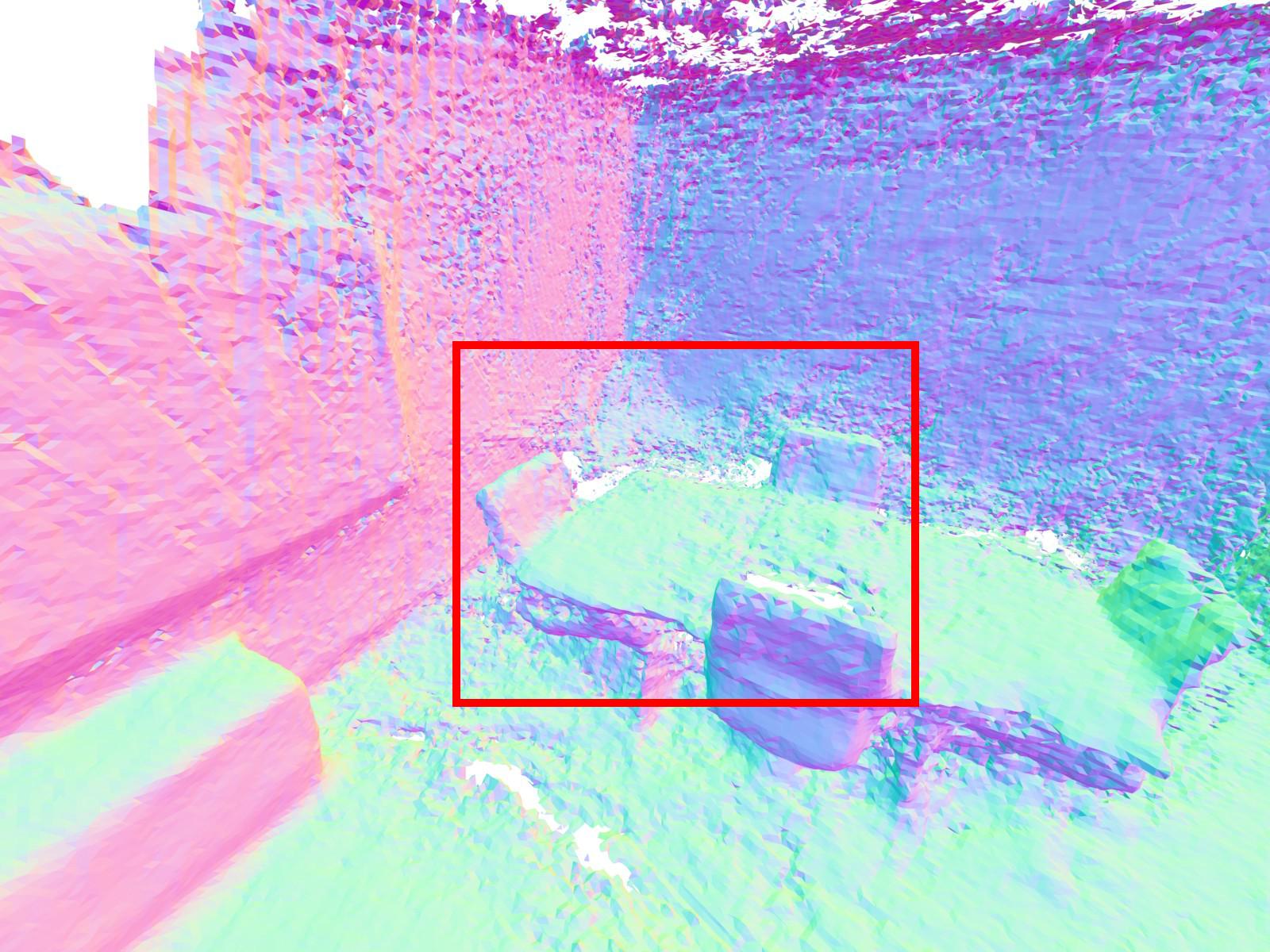} &
    \includegraphics[width=\sz\linewidth]{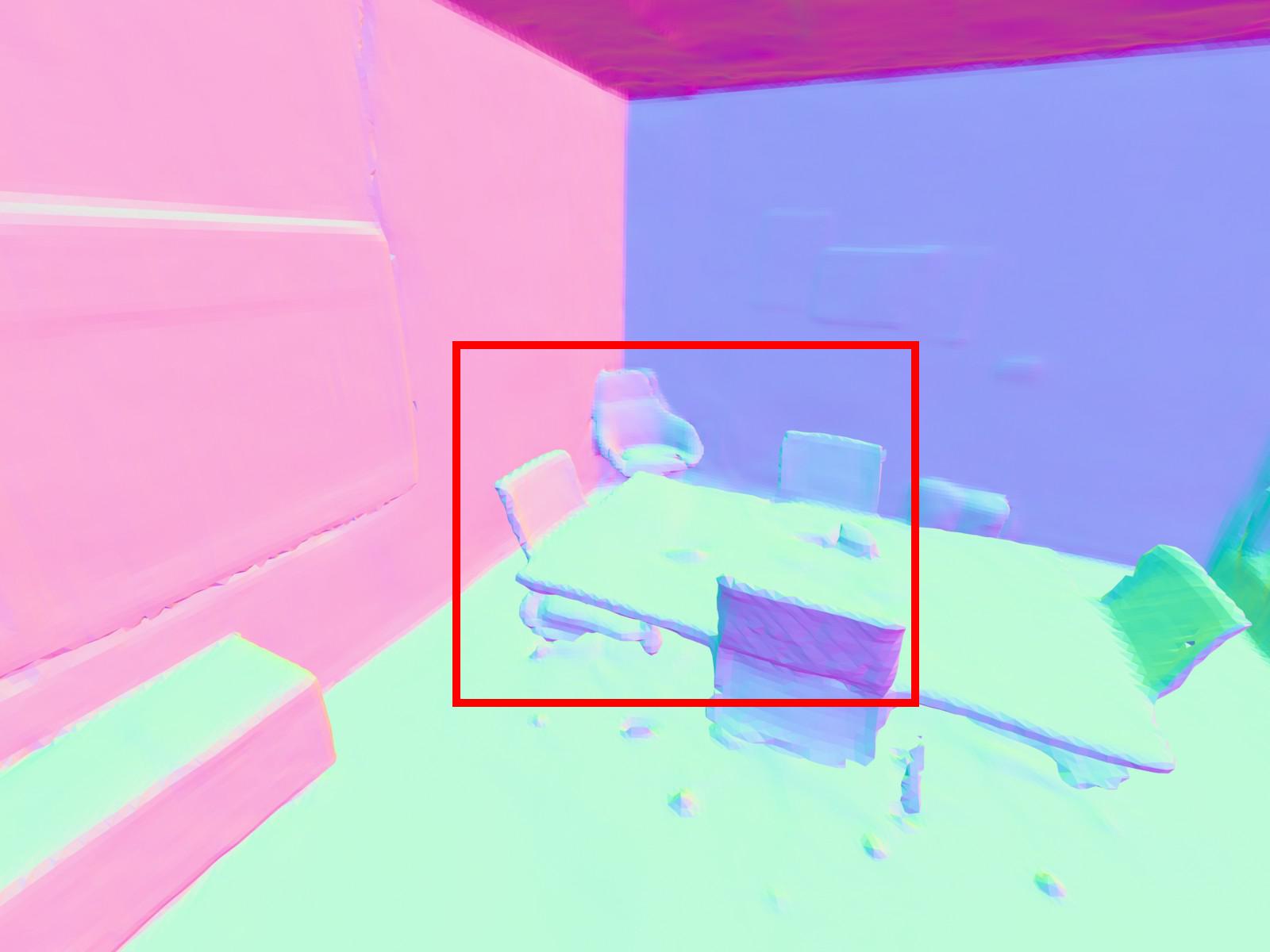} &
    \includegraphics[width=\sz\linewidth]{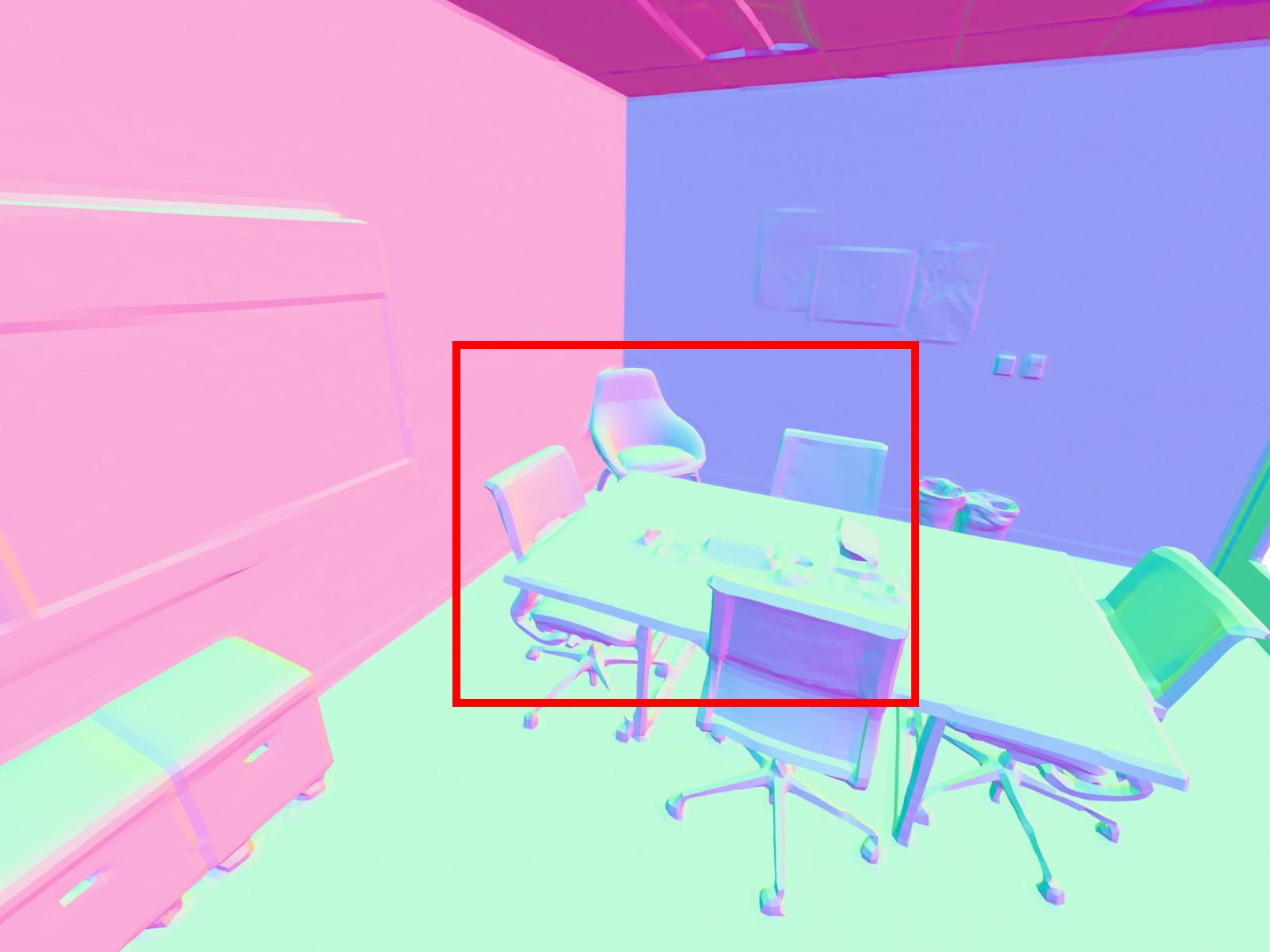} \\
    \includegraphics[width=\sz\linewidth]{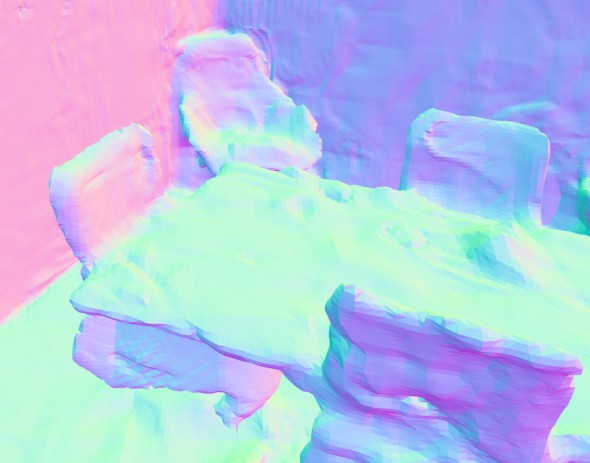} &
    \includegraphics[width=\sz\linewidth]{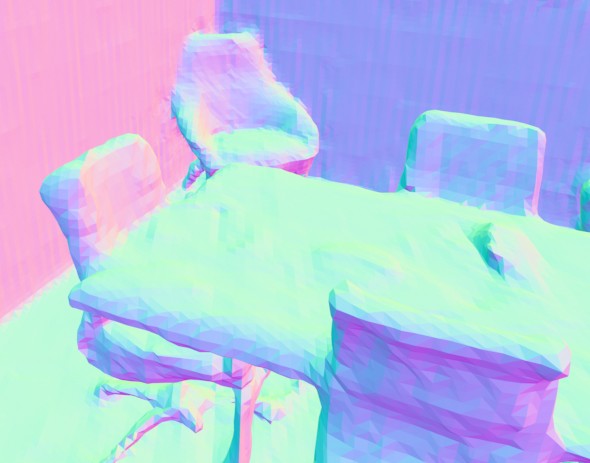} & 
    \includegraphics[width=\sz\linewidth]{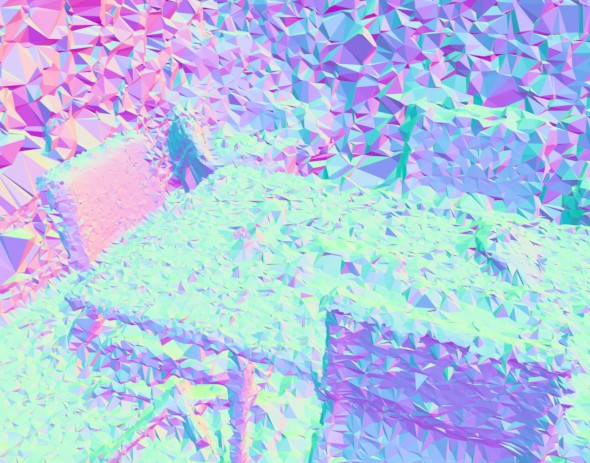} & \includegraphics[width=\sz\linewidth]{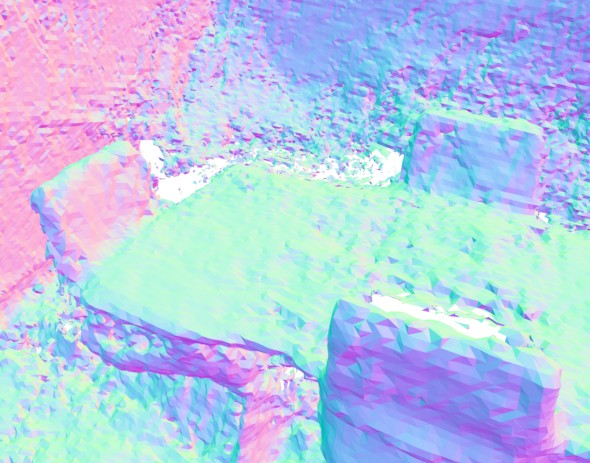} &
    \includegraphics[width=\sz\linewidth]{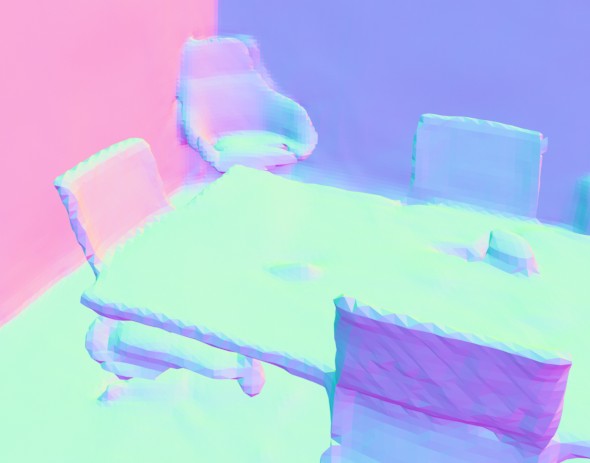} &
    \includegraphics[width=\sz\linewidth]{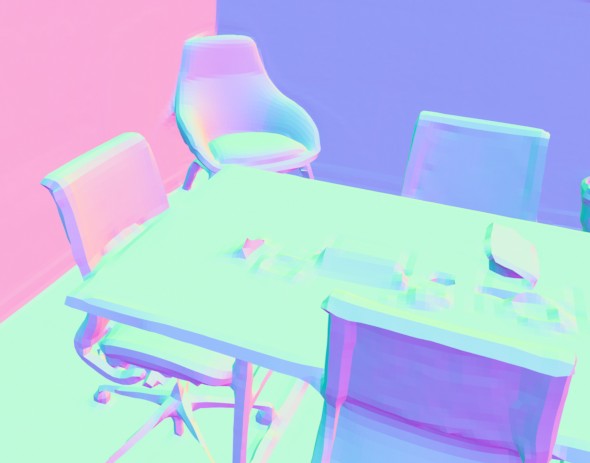} \\
    {NICE-SLAM} & {Vox-Fusion} & {COLMAP} & { DROID-SLAM} & {\textbf{NICER-SLAM}} & {GT} \\
    \multicolumn{2}{c|}{\textbf{\textit{RGB-D input}}} & \multicolumn{3}{c|}{\textbf{\textit{RGB input}}} & \\ 
  \end{tabular} 
  \caption{\textbf{3D Reconstruction Results on the Replica Dataset~\cite{replica19arxiv}.} The second and fourth row show zoom-in normal maps.
  }
  \label{fig:replica_rec}
  \vspace{-10pt}
\end{figure*}
}

\newcommand{\figreplicarendering}{
\begin{figure*}[htbp]
  \centering
  \footnotesize
  \setlength{\tabcolsep}{1.5pt}
  \newcommand{\sz}{0.162}
  \begin{tabular}{cc|ccc|c}
    \includegraphics[width=\sz\linewidth]{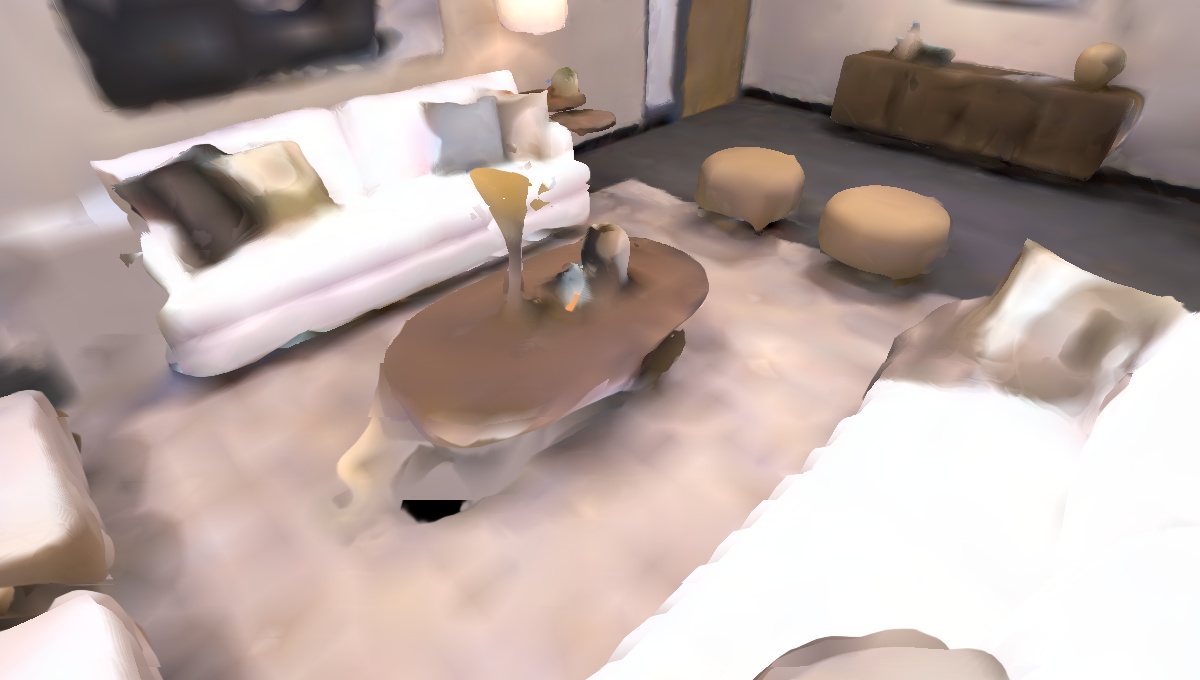} &
    \includegraphics[width=\sz\linewidth]{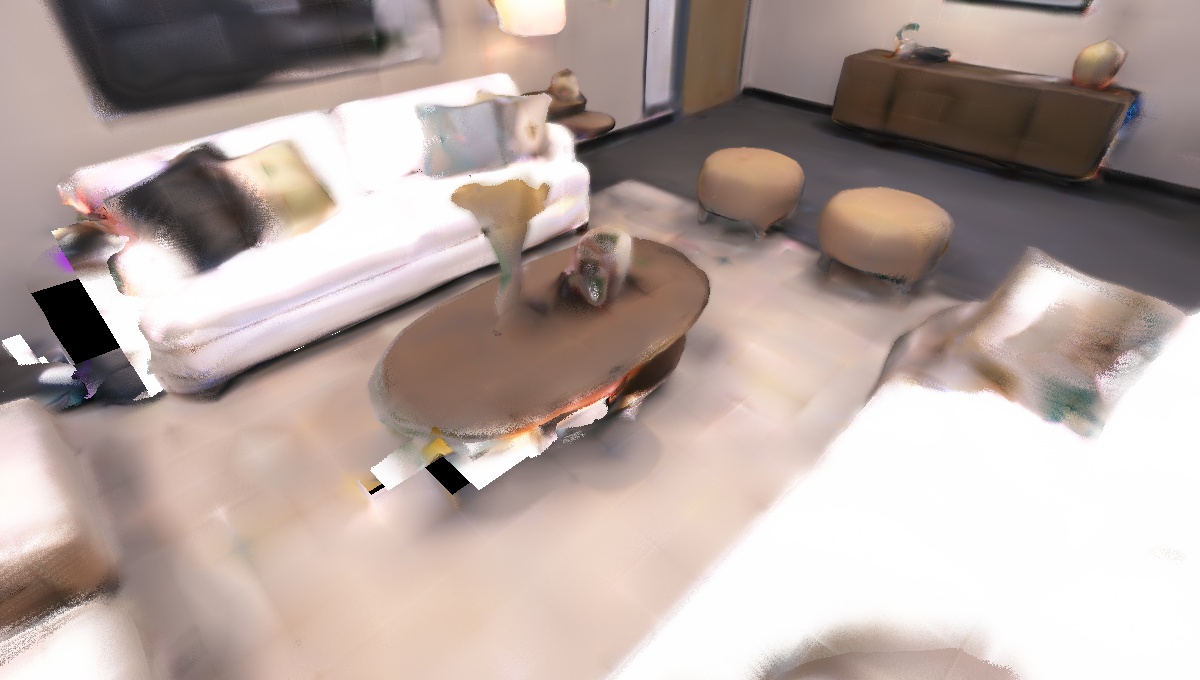} & \includegraphics[width=\sz\linewidth]{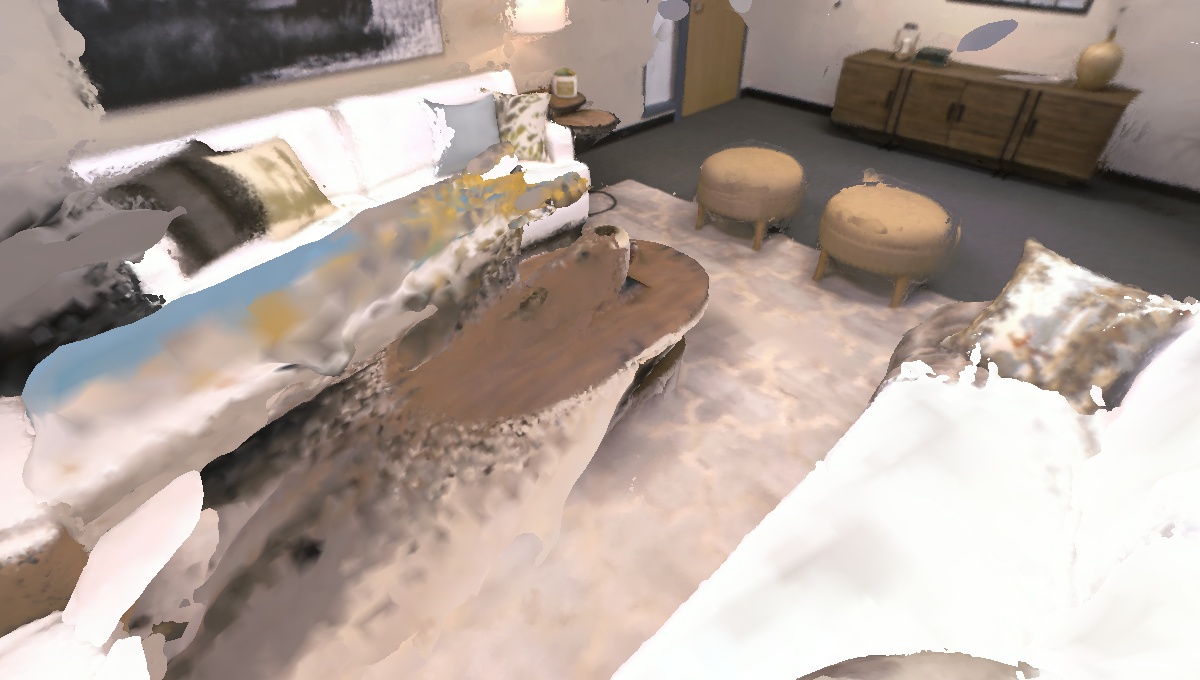} &
    \includegraphics[width=\sz\linewidth]{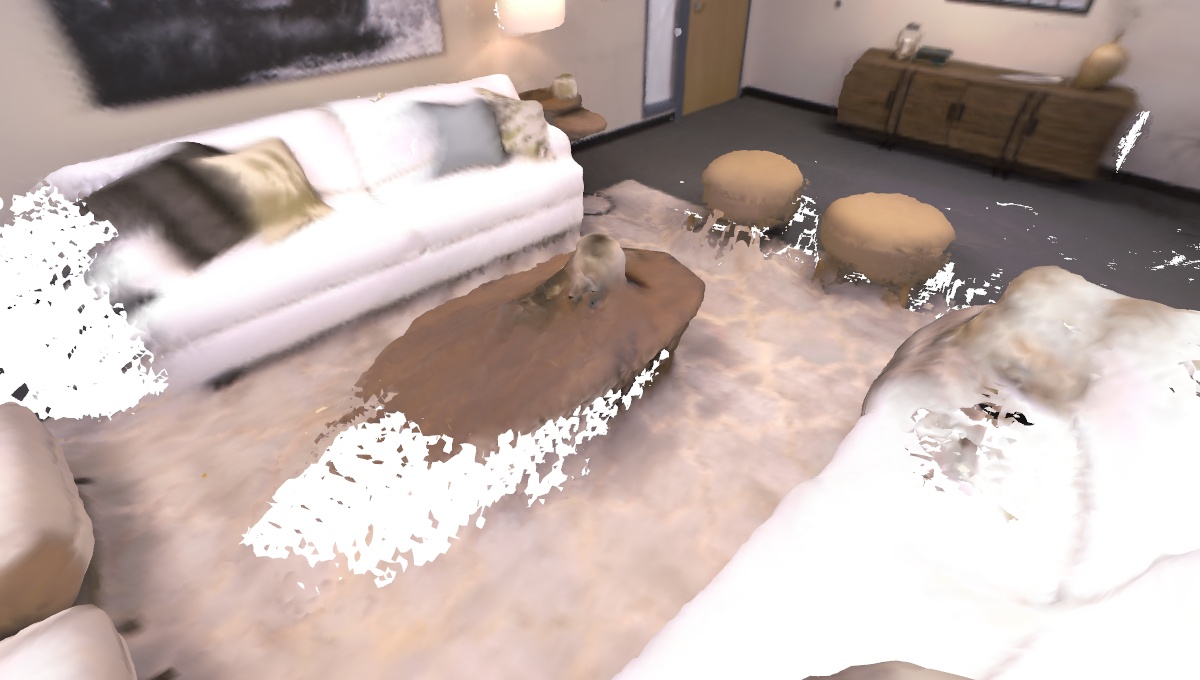} &
    \includegraphics[width=\sz\linewidth]{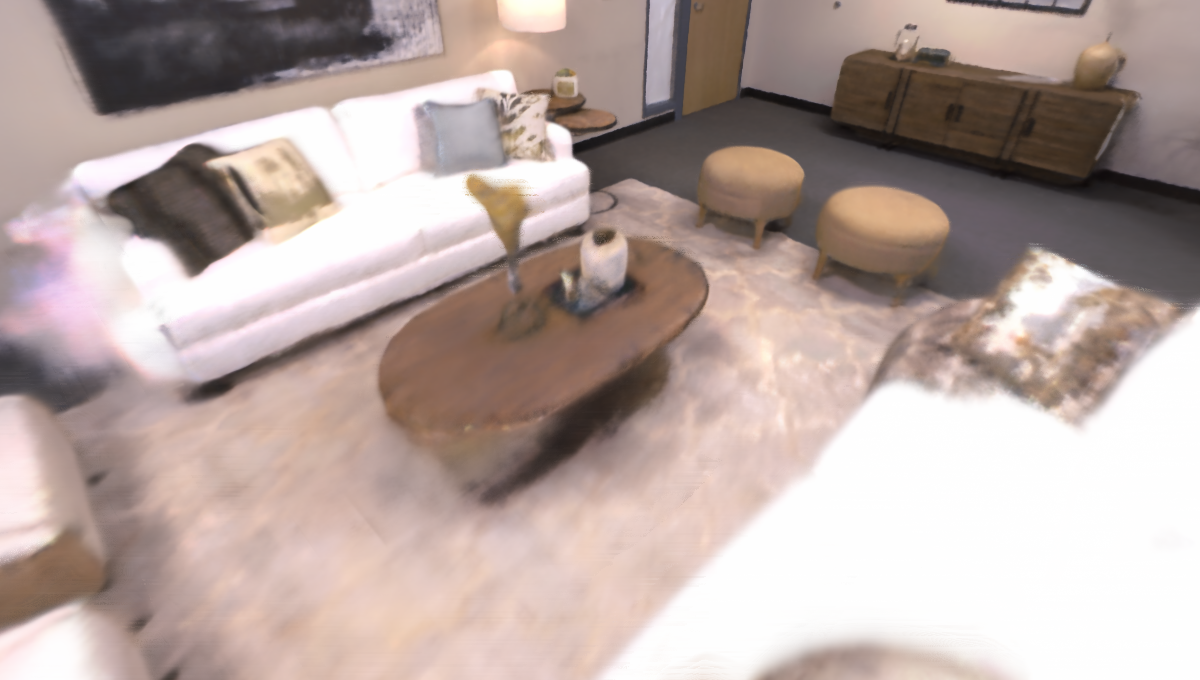} &
    \includegraphics[width=\sz\linewidth]{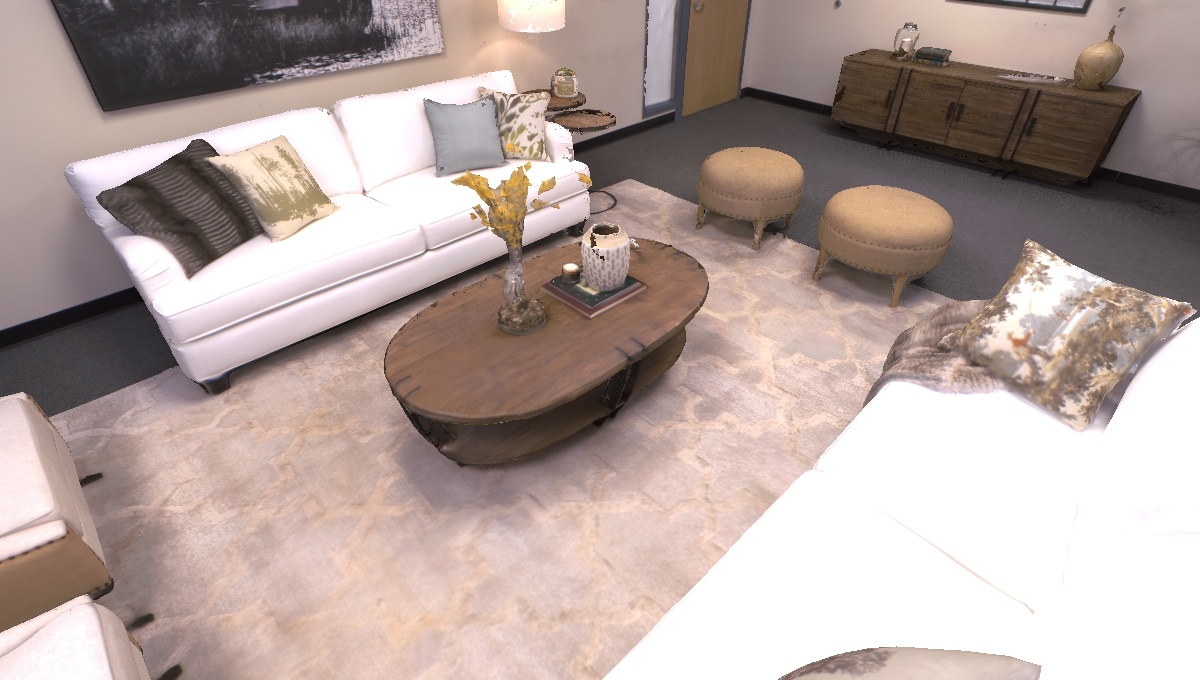}
     \\
     \includegraphics[width=\sz\linewidth]{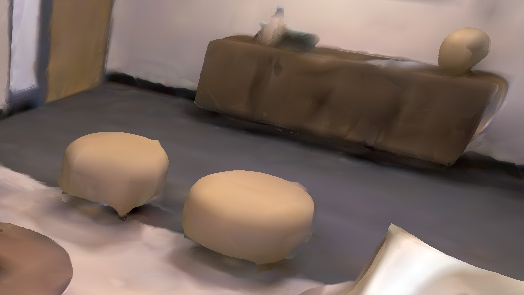} &
    \includegraphics[width=\sz\linewidth]{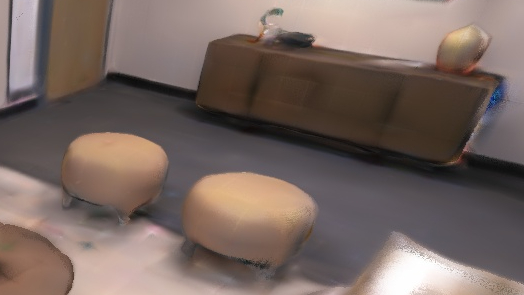} & 
    \includegraphics[width=\sz\linewidth]{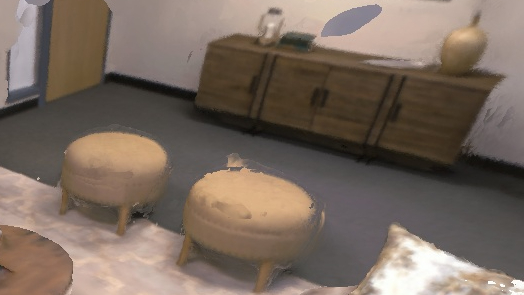} & \includegraphics[width=\sz\linewidth]{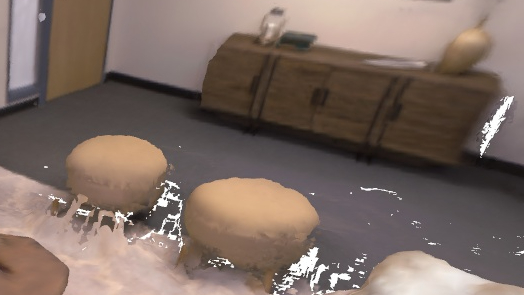} &
    \includegraphics[width=\sz\linewidth]{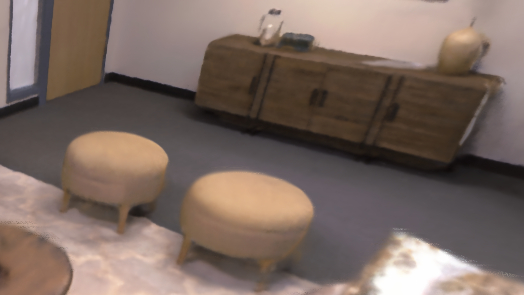} &
    \includegraphics[width=\sz\linewidth]{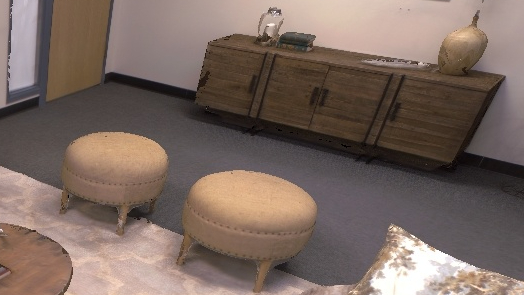} \\
    
    \includegraphics[width=\sz\linewidth]{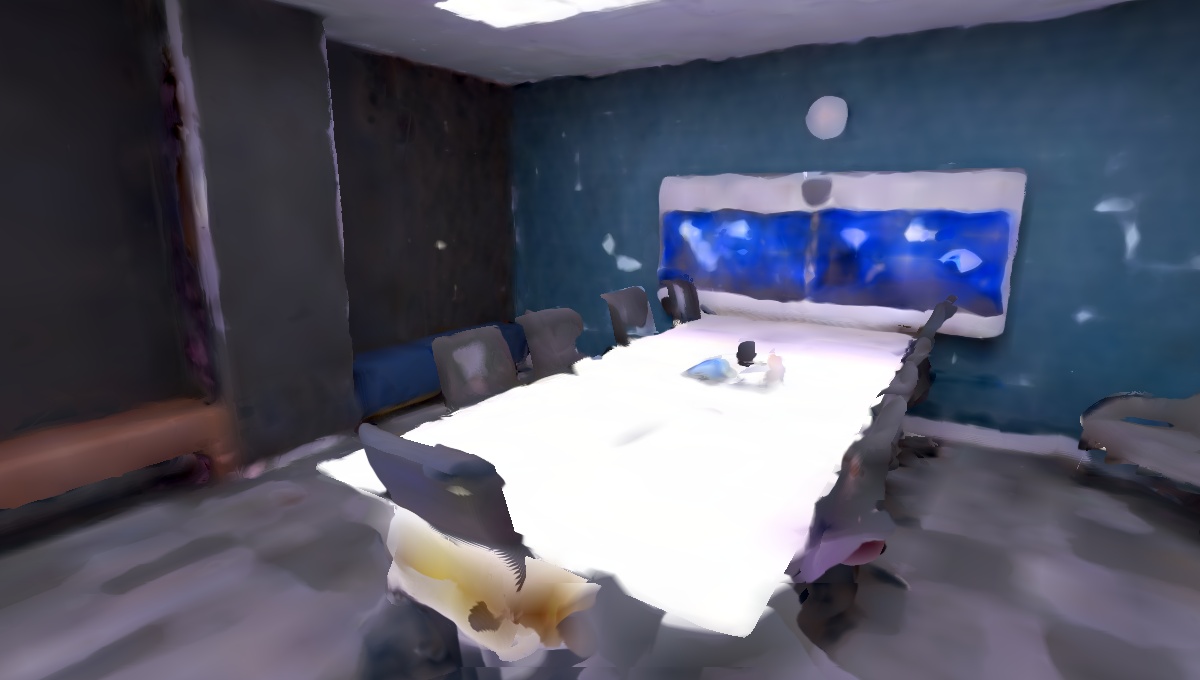} &
    \includegraphics[width=\sz\linewidth]{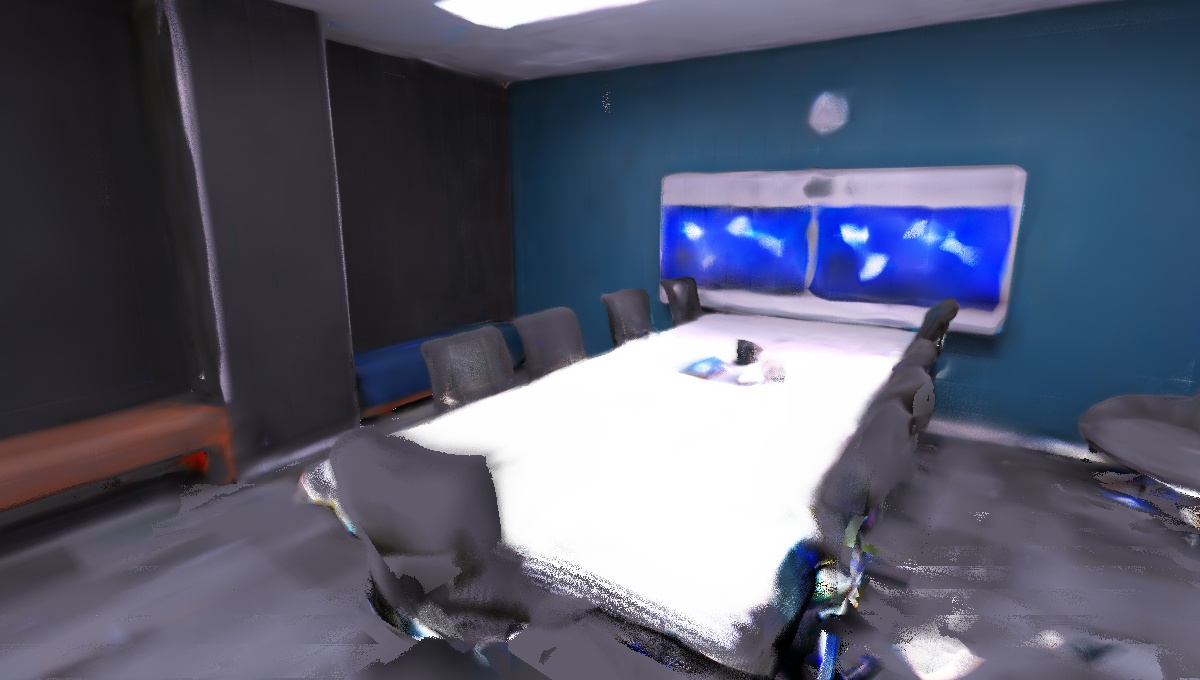} & \includegraphics[width=\sz\linewidth]{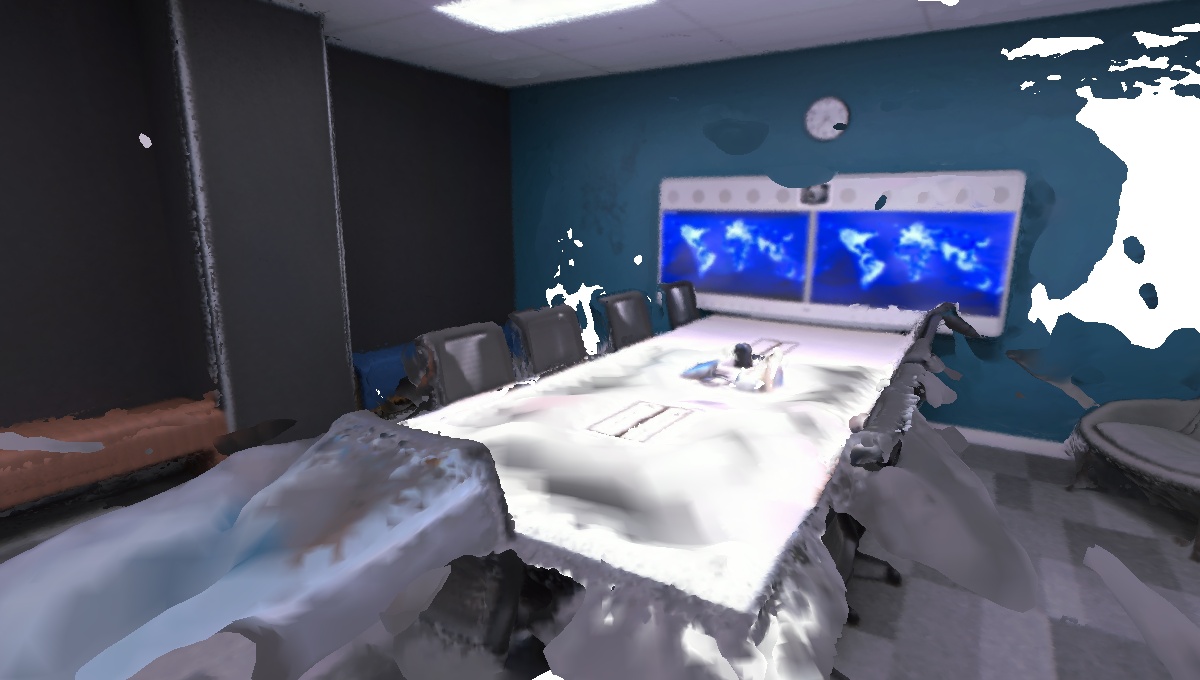} &
    \includegraphics[width=\sz\linewidth]{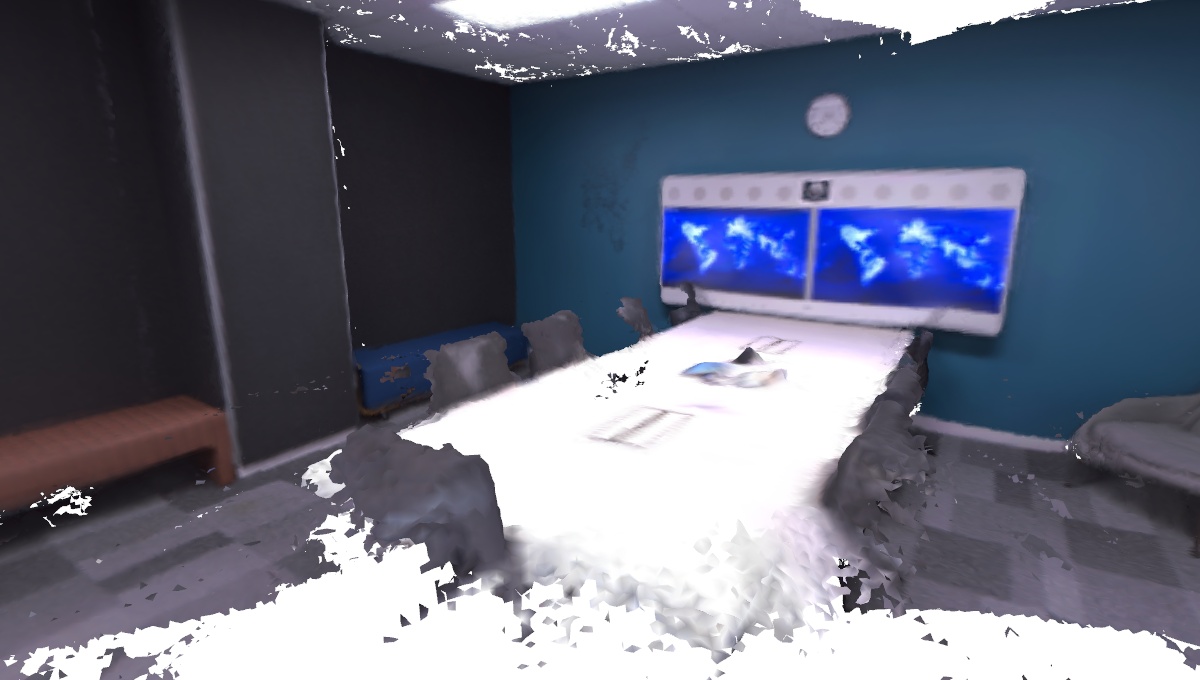} &
    \includegraphics[width=\sz\linewidth]{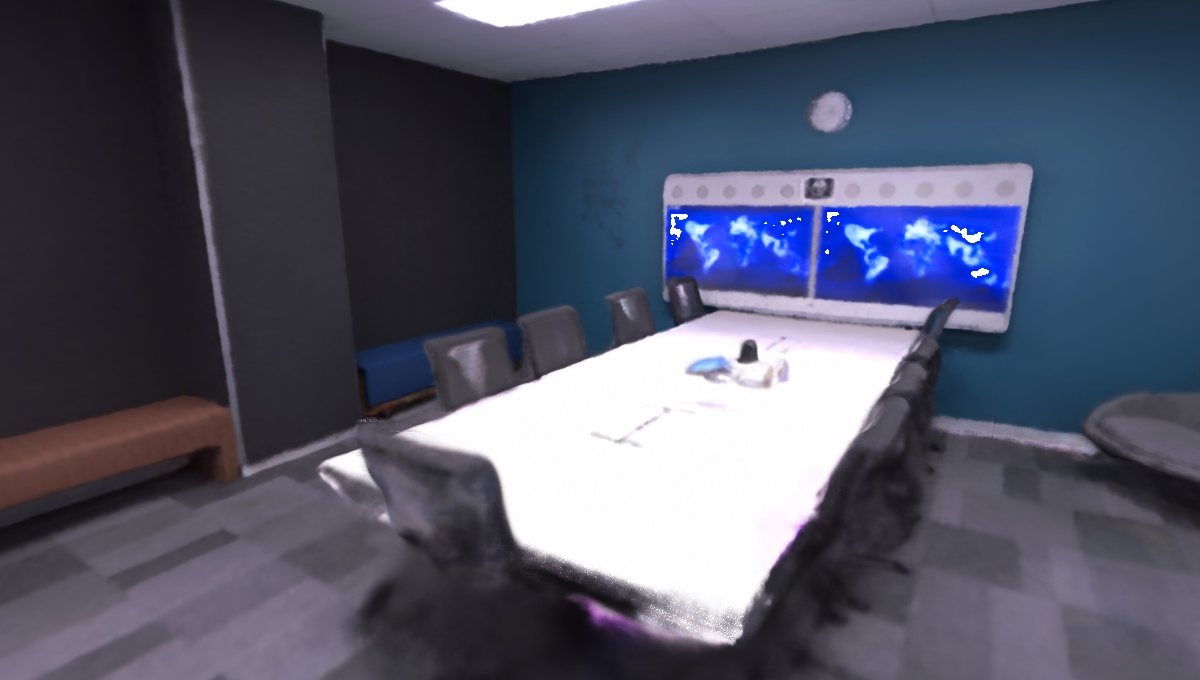} &
    \includegraphics[width=\sz\linewidth]{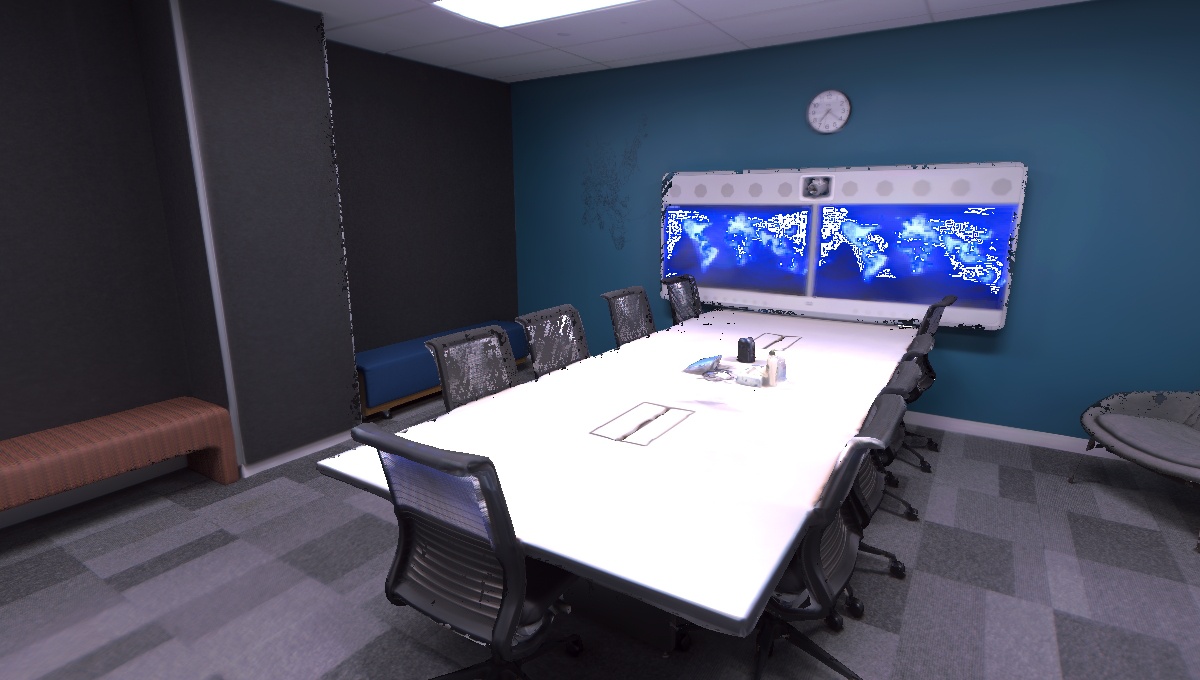}
     \\
     \includegraphics[width=\sz\linewidth]{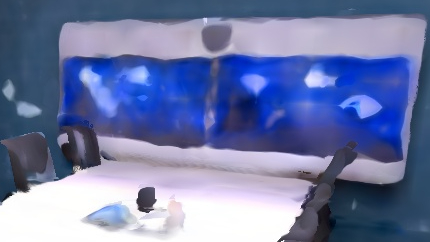} &
    \includegraphics[width=\sz\linewidth]{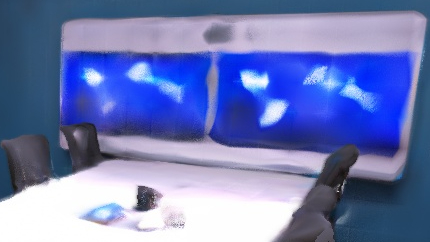} & 
    \includegraphics[width=\sz\linewidth]{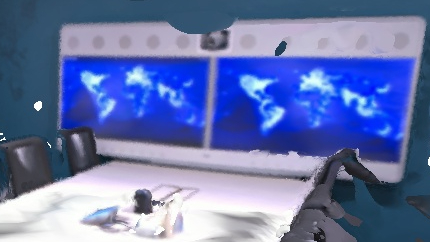} & \includegraphics[width=\sz\linewidth]{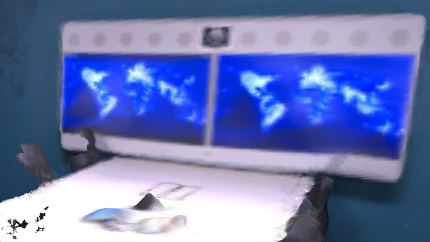} &
    \includegraphics[width=\sz\linewidth]{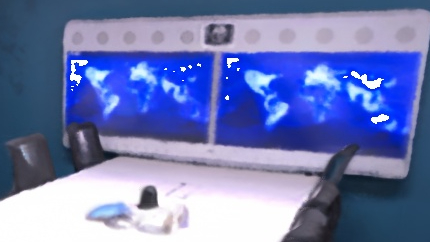} &
    \includegraphics[width=\sz\linewidth]{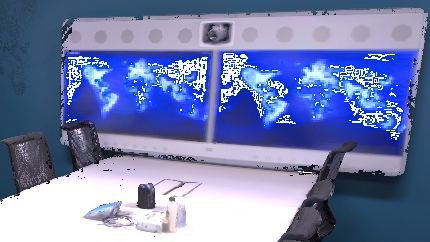} \\

    {NICE-SLAM} & {Vox-Fusion} & {COLMAP} & { DROID-SLAM} & {\textbf{NICER-SLAM}} & {GT} \\
    \multicolumn{2}{c|}{\textbf{\textit{RGB-D input}}} & \multicolumn{3}{c|}{\textbf{\textit{RGB input}}} & \\ 
  \end{tabular} 
  \caption{\textbf{Novel View Synthesis Results on the Replica the Dataset~\cite{replica19arxiv}.} 
  The second and fourth row show zoom-in renderings for better comparison.
  Note that we selected novel viewpoints far from the input views (extrapolation).
}
  \label{fig:replica_rendering}
\end{figure*}
}

\newcommand{\figsevenscenesreconstruction}{
\begin{figure*}[htb]
  \centering
  \footnotesize
  \setlength{\tabcolsep}{1.5pt}
  \newcommand{\sz}{0.162}
  \begin{tabular}{cc|ccc|c}
    \includegraphics[width=\sz\linewidth]{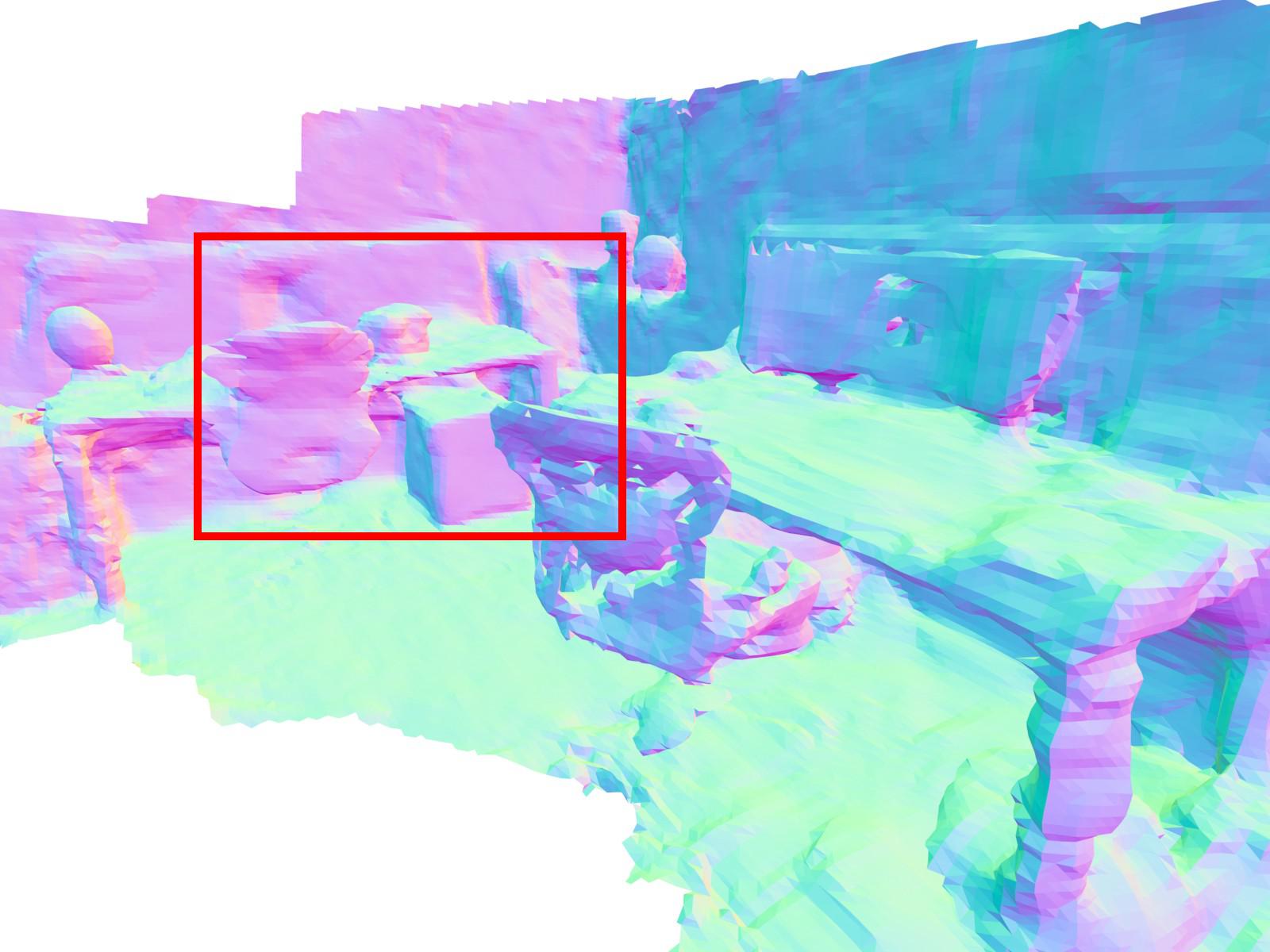} &
    \includegraphics[width=\sz\linewidth]{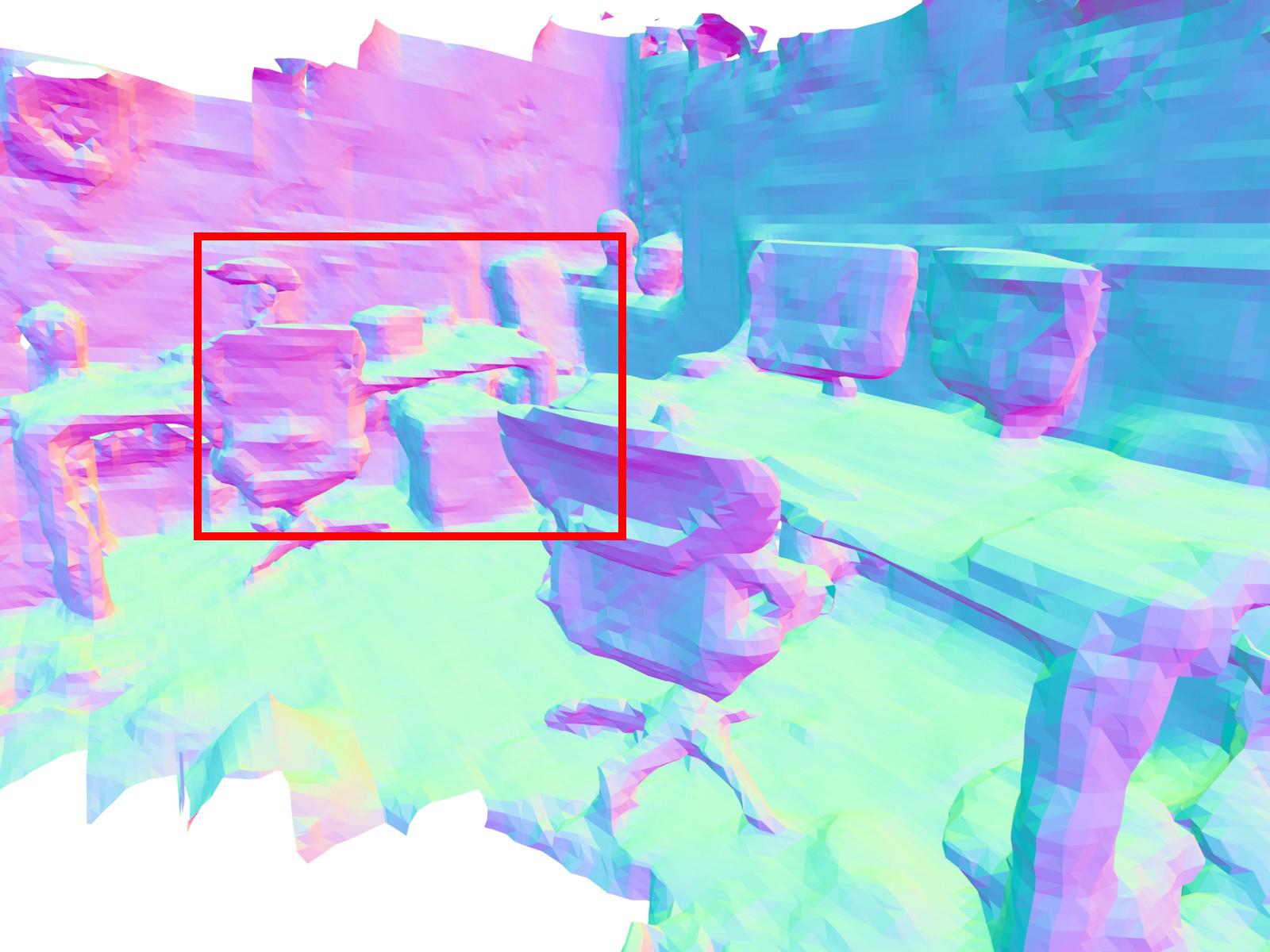} & \includegraphics[width=\sz\linewidth]{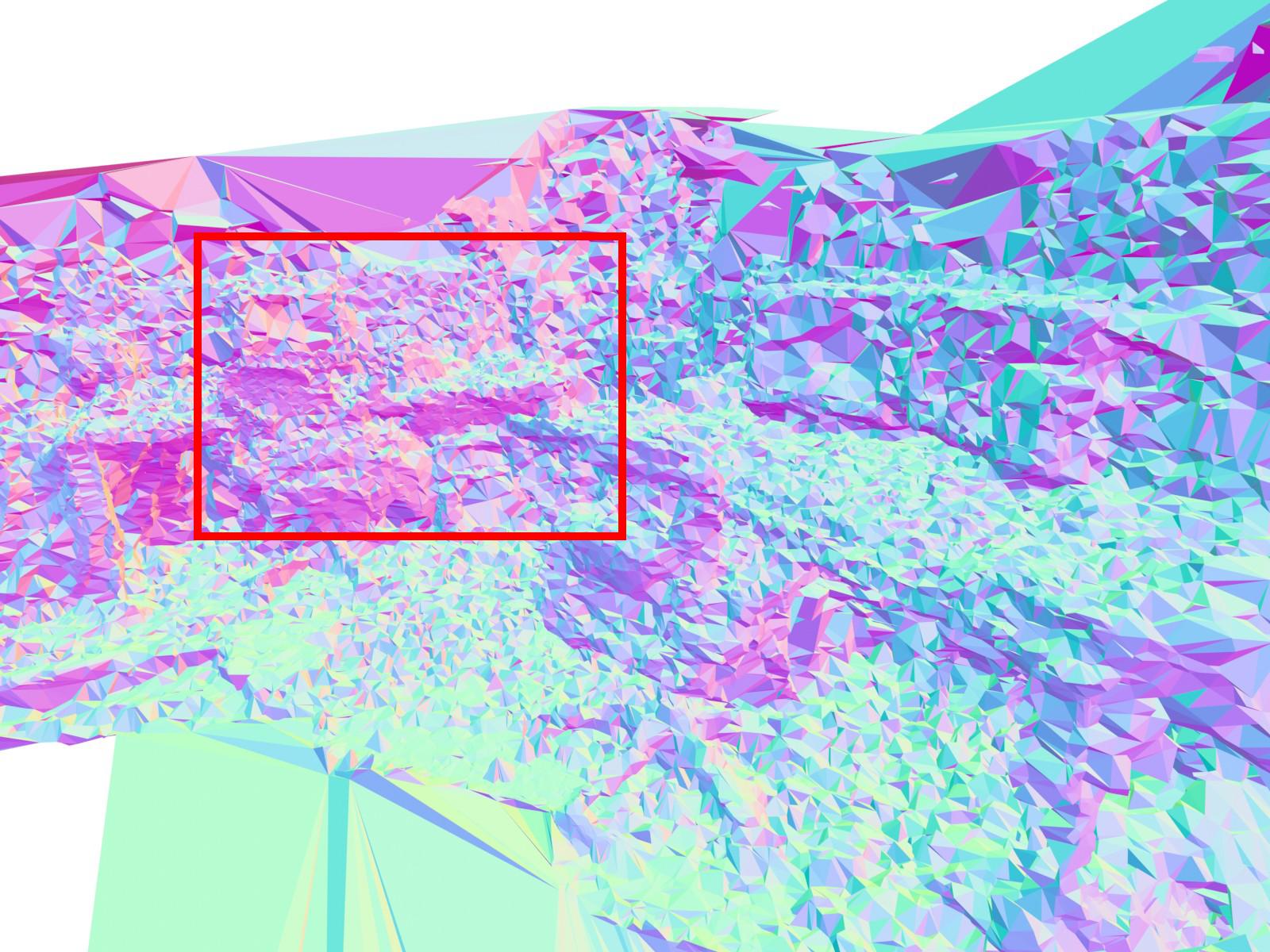} &
    \includegraphics[width=\sz\linewidth]{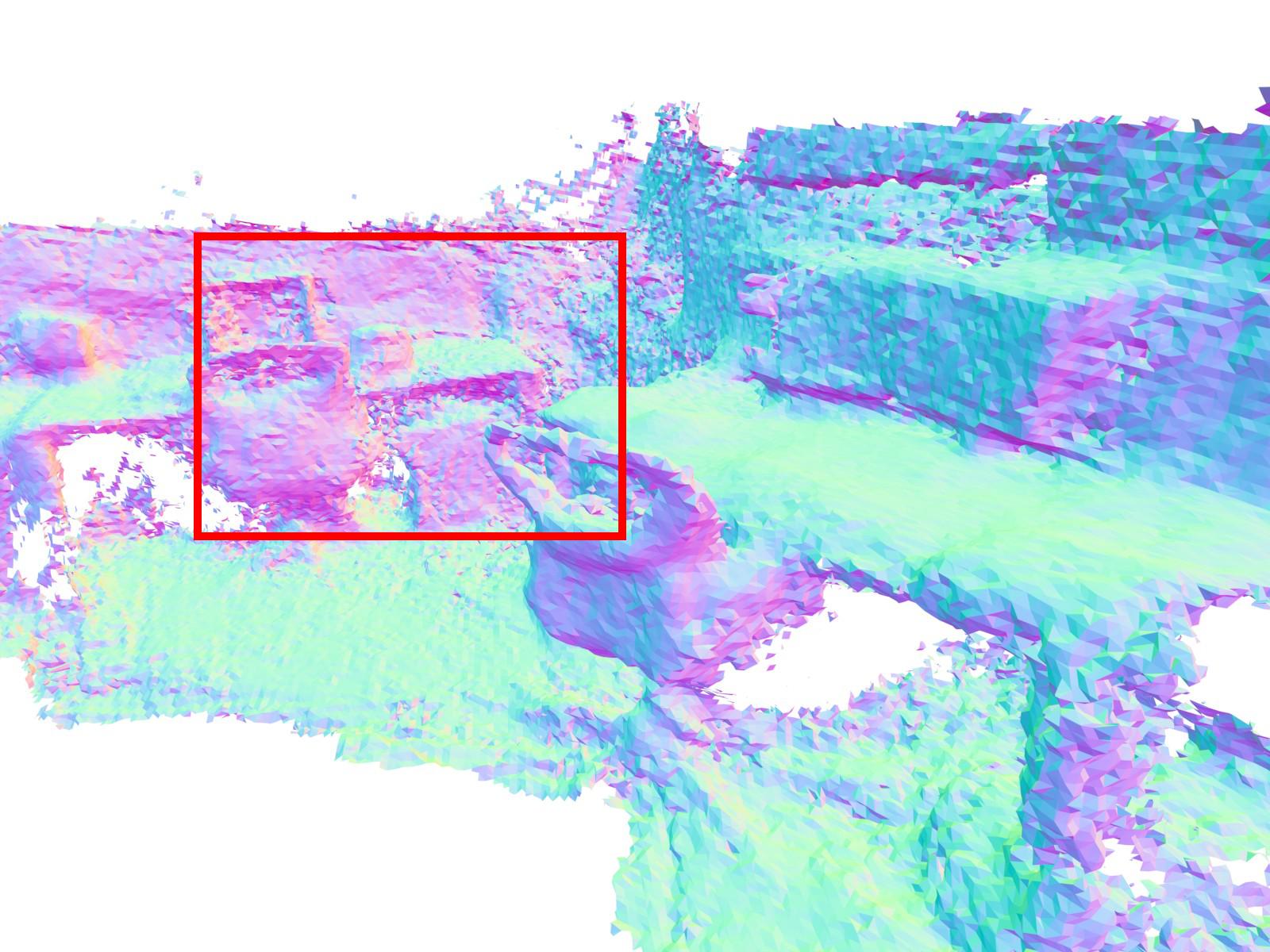} &
    \includegraphics[width=\sz\linewidth]{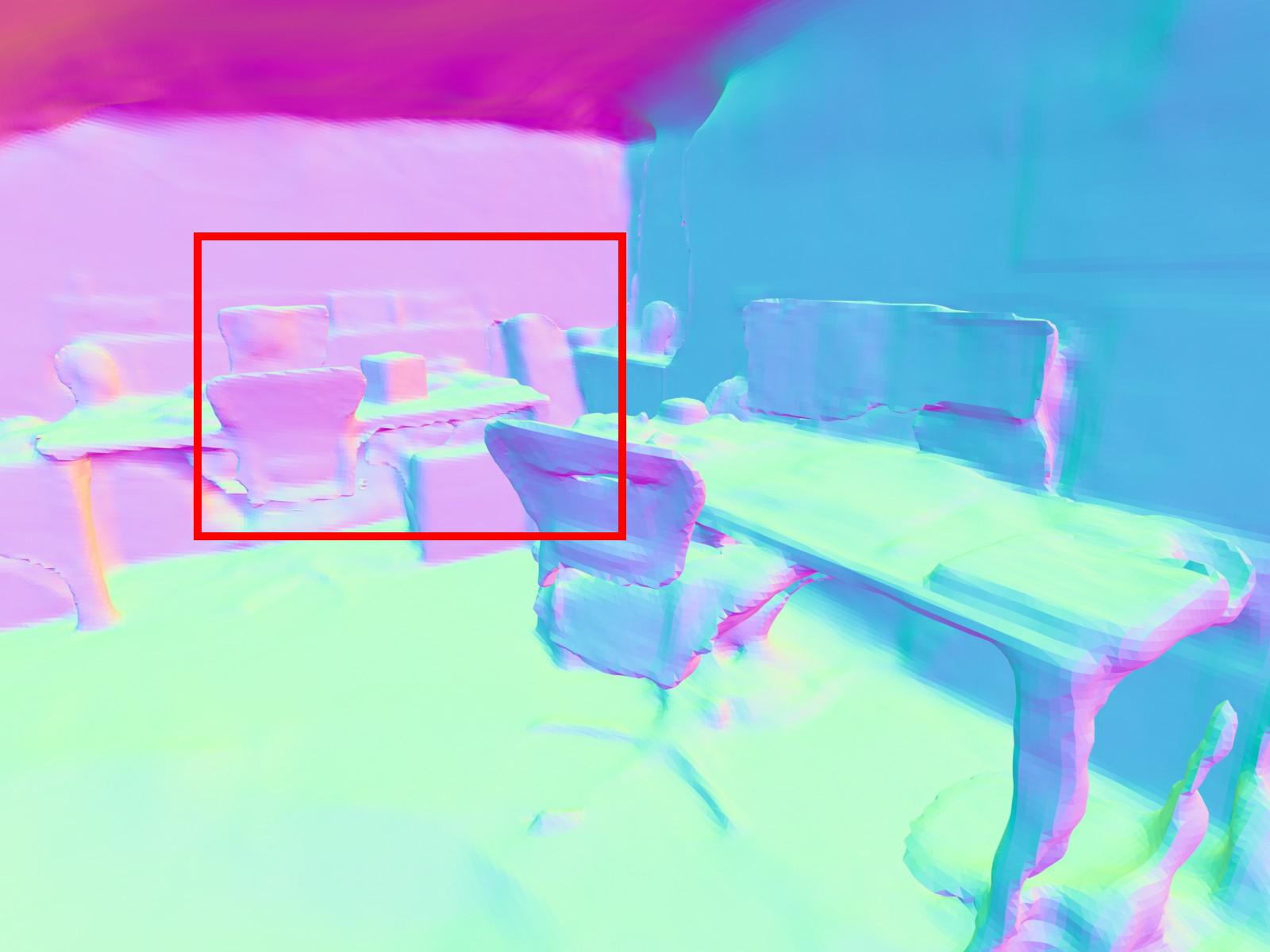} &
    \includegraphics[width=\sz\linewidth]{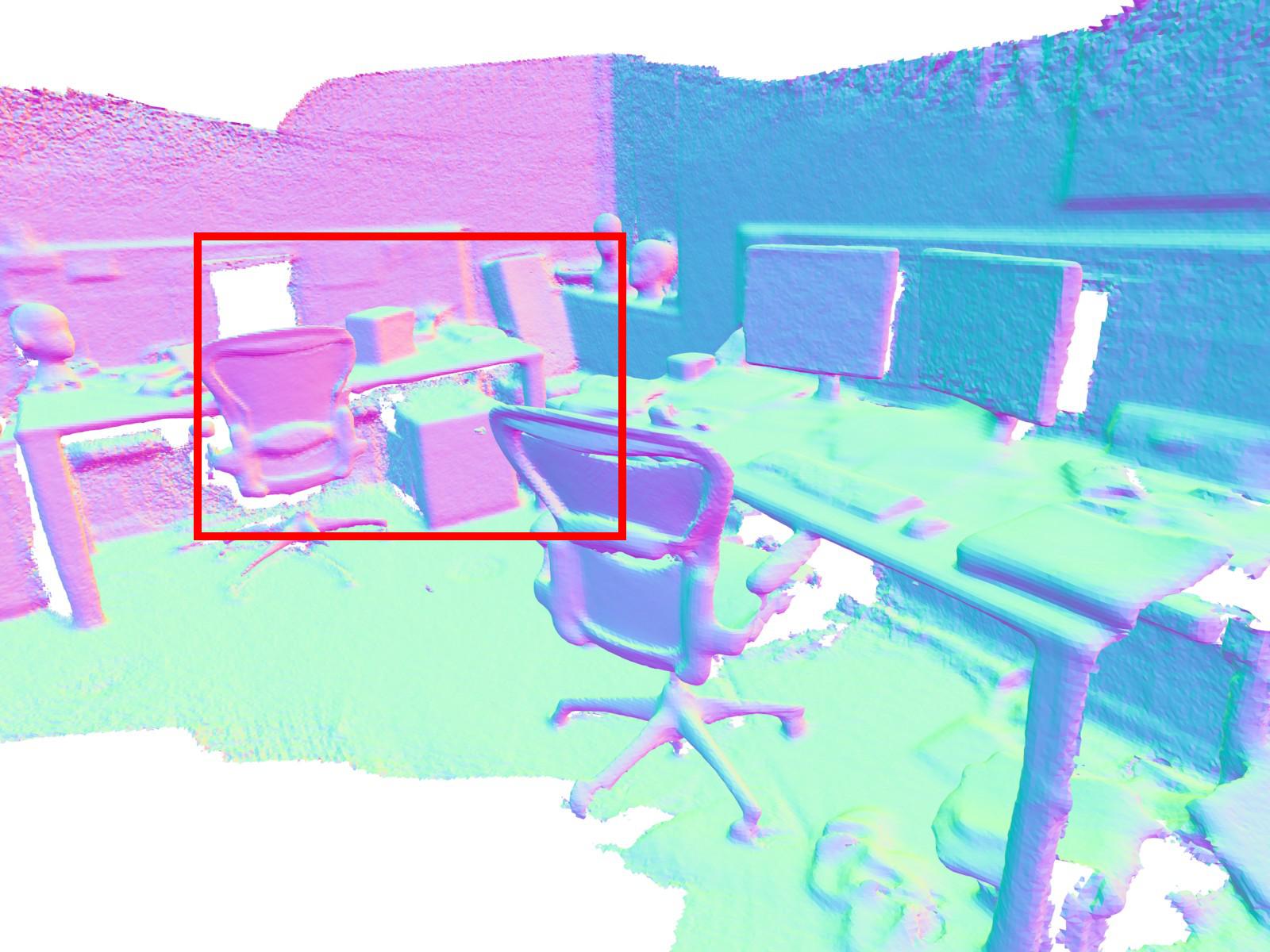}
     \\
     \includegraphics[width=\sz\linewidth]{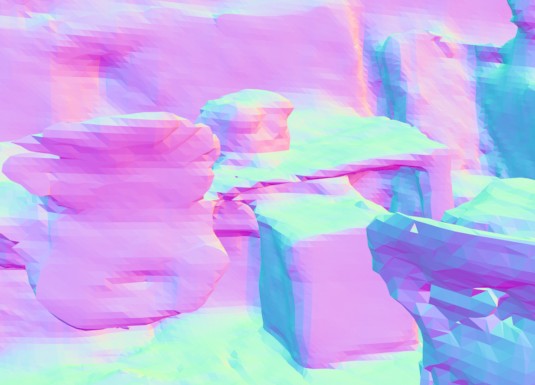} &
    \includegraphics[width=\sz\linewidth]{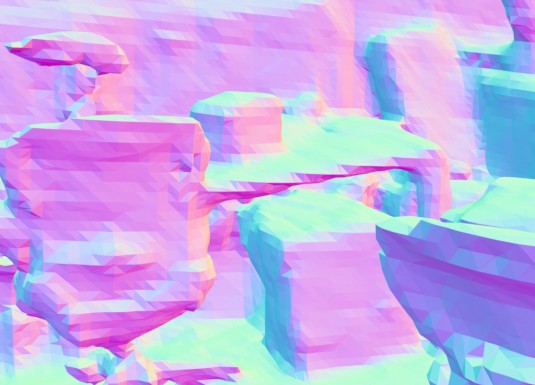} & 
    \includegraphics[width=\sz\linewidth]{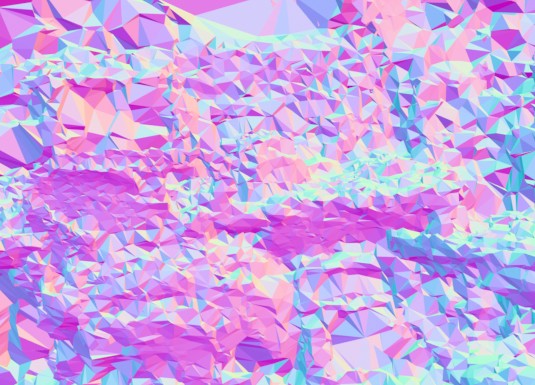} & \includegraphics[width=\sz\linewidth]{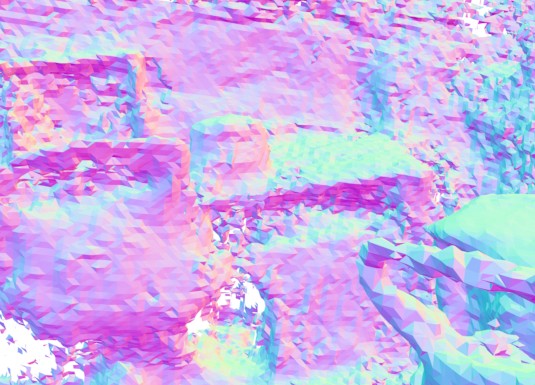} &
    \includegraphics[width=\sz\linewidth]{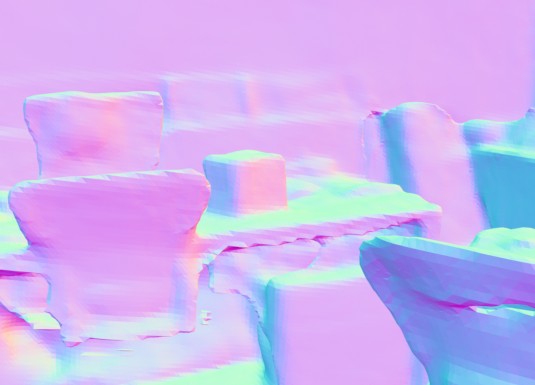} &
    \includegraphics[width=\sz\linewidth]{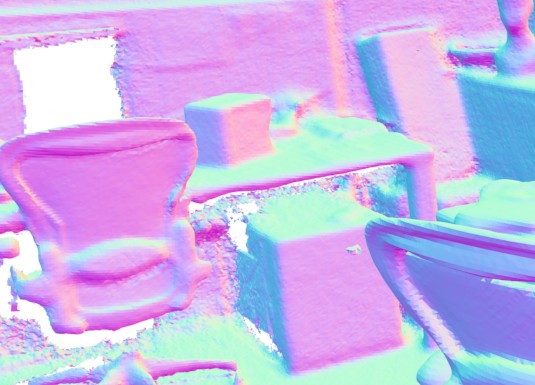} \\
    
    {NICE-SLAM} & {Vox-Fusion} & {COLMAP} & { DROID-SLAM} & {\textbf{NICER-SLAM}} & {GT} \\
    \multicolumn{2}{c|}{\textbf{\textit{RGB-D input}}} & \multicolumn{3}{c|}{\textbf{\textit{RGB input}}} & \\ 
  \end{tabular} 
  \caption{\textbf{3D Reconstruction Results on the 7-Scenes Dataset~\cite{7scenes}.} The second row shows zoomed-in normal maps. It is apparent that the quality of the scene representation is substantially worse for the other RGB-based methods despite their better tracking accuracy.
  }
  \label{fig:7scenes}
\end{figure*}
}

\newcommand{\figablationrepresentation}{
\begin{figure}[htbp]
  \centering
  \setlength{\tabcolsep}{1.5pt}
  \newcommand{\sz}{0.32}
  \begin{tabular}{ccc}
    \includegraphics[width=\sz\linewidth]{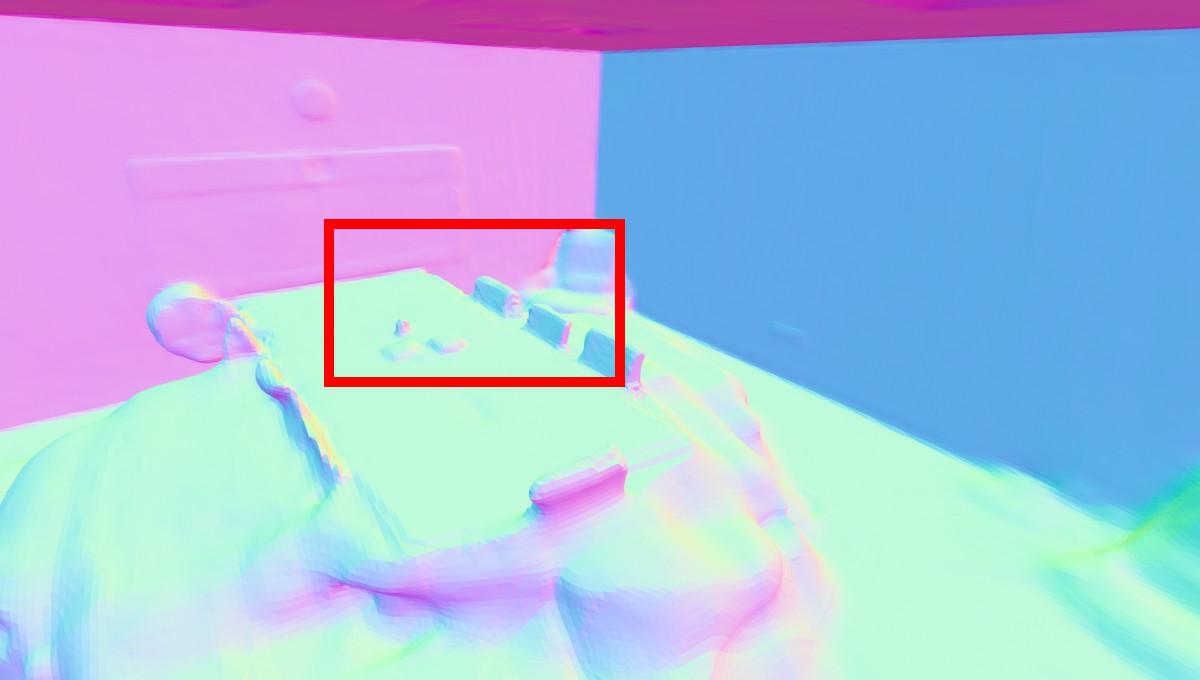} &
    \includegraphics[width=\sz\linewidth]{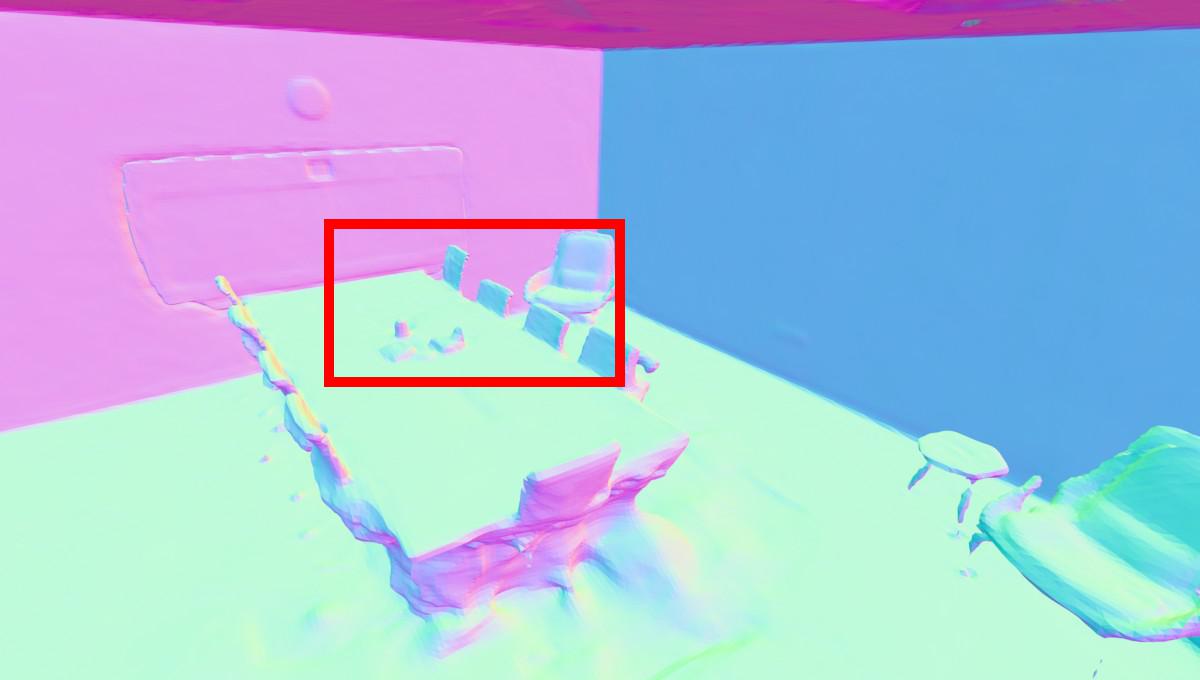} &
    \includegraphics[width=\sz\linewidth]{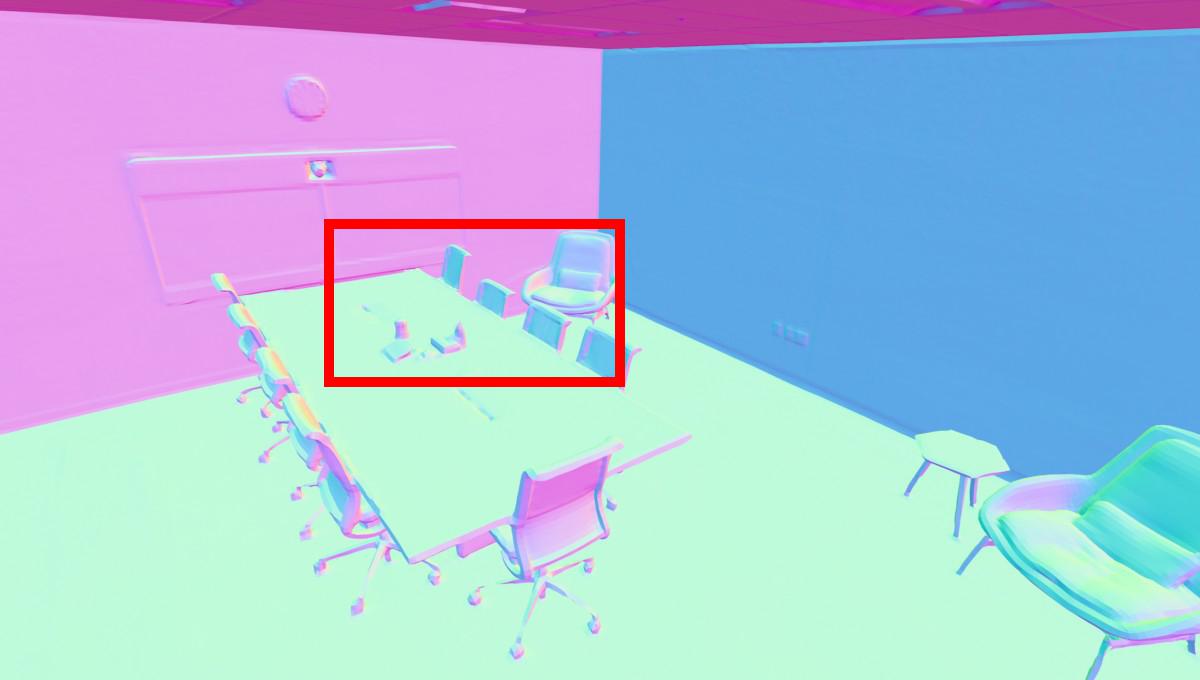} \\
    \includegraphics[width=\sz\linewidth]{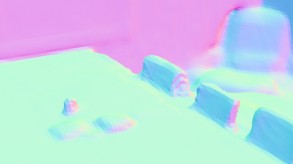} &
    \includegraphics[width=\sz\linewidth]{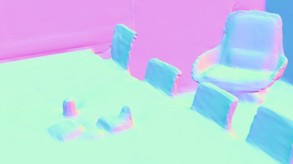} &
    \includegraphics[width=\sz\linewidth]{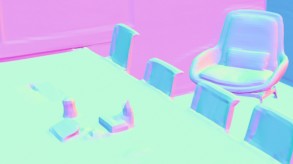} \\

    {w/ Occupancy} & {w/ SDF} & {GT} \\
  \end{tabular} 
  \caption{\textbf{Ablation Study on SDF vs. Occupancy.} We conduct the ablation on one random Replica scene (\texttt{office-4}). The second row depicts zoomed-in normal maps for better comparison.
  }
  \label{fig:ablation_representation}
  \vspace{-6pt}
\end{figure}
}

\colorlet{colorFst}{Green!25}       %
\colorlet{colorSnd}{SpringGreen!45} %
\colorlet{colorTrd}{Yellow!30}      %

\colorlet{colorSep}{blue!5}         %

\newcommand{\fs}{\cellcolor{colorFst}\bf}   %
\newcommand{\nd}{\cellcolor{colorSnd}}      %
\newcommand{\rd}{\cellcolor{colorTrd}}      %

\newcommand{\tablereplicatracking}{
\begin{table}[htbp]
  \centering
  \footnotesize
  \setlength{\tabcolsep}{3.5pt}          %
  \renewcommand{\arraystretch}{1.05}     %
  \resizebox{\linewidth}{!}{
  \begin{tabular}{@{\hspace*{1em}}llcccccccccccc}
    \toprule
    & \multicolumn{1}{c}{\makecell{\tt{rm-0}}} & \multicolumn{1}{c}{\makecell{\tt{rm-1}}} &  \multicolumn{1}{c}{\makecell{\tt{rm-2}}} & \multicolumn{1}{c}{\makecell{\tt{off-0}}} & \multicolumn{1}{c}{\makecell{\tt{off-1}}} & \multicolumn{1}{c}{\makecell{\tt{off-2}}}& \multicolumn{1}{c}{\makecell{\tt{off-3}}} & \multicolumn{1}{c}{\makecell{\tt{off-4}}} & Avg. \\
    \midrule
    \rowcolor{colorSep}
    \multicolumn{12}{l}{\textit{\textbf{RGB-D input}}}\\
    NICE-SLAM  & 1.69 & 2.04 & 1.55 & 0.99 & 0.90 & 1.39 & 3.97 & 3.08 & 1.95 \\
    Vox-Fusion & \fs 0.27 & \rd 1.33 & 0.47 & \rd 0.70 & 1.11 & \nd 0.46 & \nd 0.26 & \nd 0.58 & \rd 0.65 \\    
    \midrule
    \rowcolor{colorSep}
    \multicolumn{12}{l}{\textit{\textbf{RGB input}}}\\
    COLMAP     & 0.62 & 23.7 & \rd 0.39 & \nd 0.33 & \fs 0.24 & 0.79 & \fs 0.14 & 1.73 & 3.49 \\
    DROID-SLAM & \nd 0.34 & \fs 0.13 & \fs 0.27 & \fs 0.25 & \rd 0.42 & \fs 0.32 & \rd 0.52 & \fs 0.40 & \fs \textbf{0.33} \\
    DROID-SLAM$^*$ & \rd 0.58 & \nd 0.58 & \nd 0.38 & 1.06 & \nd 0.40 & \rd 0.70 & 0.53 & \rd 1.33 & \rd 0.70 \\
    \textbf{NICER-SLAM} & 1.36 & 1.60 & 1.14 & 2.12 & 3.23 & 2.12 & 1.42 & 2.01 & 1.88 \\
    \bottomrule
  \end{tabular}}%
  \caption{\textbf{Camera Tracking Results on the Replica Dataset~\cite{replica19arxiv}.} ATE RMSE [cm] ($\downarrow$) is used as the evaluation metric.}
  \label{tab:replica_tracking}
  \vspace{-5pt}
\end{table}
}

\newcommand{\tablesevenscenestracking}{
\begin{table}[htbp]
  \centering
  \footnotesize
  \setlength{\tabcolsep}{1.6pt}          %
  \renewcommand{\arraystretch}{1.05}     %
  \resizebox{\linewidth}{!}{
  \begin{tabular}{@{\hspace*{1em}}llcccccccccccc}
    \toprule
    &  \tt{chess} &  \tt{fire} &  \tt{heads} & \tt{office} &  \tt{pumpkin} & \tt{kitchen} & \tt{stairs} & \tt{Avg.} \\
    \midrule
    \rowcolor{colorSep}
    \multicolumn{12}{l}{\textit{\textbf{RGB-D input}}}\\
    NICE-SLAM      & \fs 2.16 & \fs 1.63 & 7.80 & \nd 5.73 & \rd 19.34 & \nd 3.31 & 4.31 & \rd 6.33 \\
    Vox-Fusion     & \nd 2.53 & \nd 1.91 & \rd 1.94 & \fs 5.26 & \fs 15.33 & \fs 2.79 & 3.40 & \fs 4.74 \\    
    \midrule
    \rowcolor{colorSep}
    \multicolumn{12}{l}{\textit{\textbf{RGB input}}}\\
    COLMAP         & 3.42 & \rd 2.40 & \nd 1.70 & 10.42 & 52.32 & 5.08 & \rd 2.62 & 11.14 \\
    DROID-SLAM     & 3.36 & \rd 2.40 & \fs 1.43 & \rd 9.19 & \nd 16.46 & 4.94 & \fs 1.85 & \nd 5.66 \\
    DROID-SLAM$^*$ & 3.55 & 2.49 & 3.32 & 10.93 & 48.53 & 4.72 & \nd 2.58 & 10.87 \\
    \textbf{NICER-SLAM}     & \rd 3.28 & 6.85 & 4.16 & 10.84 & 20.00 & \rd 3.94 & 10.81 &  8.55  \\
    \bottomrule
  \end{tabular}}
  \caption{\textbf{Camera Tracking Results on the 7-Scenes Dataset~\cite{7scenes}.} ATE RMSE [cm] ($\downarrow$) is used as the evaluation metric.}
  \label{tab:7scenes_tracking}
\end{table}
}

\newcommand{\tablereplicarendering}{
\begin{table}[htbp]
  \centering
  \footnotesize
  \setlength{\tabcolsep}{2.2pt}          %
  \renewcommand{\arraystretch}{1.1}      %
  \resizebox{\linewidth}{!}{
    \begin{tabular}{lllccccccccc}
      \toprule
         & & & \tt{rm-0} &  \tt{rm-1} &  \tt{rm-2} & \tt{off-0} &  \tt{off-1} & \tt{off-2} & \tt{off-3} & \tt{off-4} & Avg. \\    
      \midrule
      \rowcolor{colorSep}
      \multicolumn{12}{l}{\textit{\textbf{RGB-D input}}}\\
      \multirow{6}{*}{\rotatebox[origin=c]{90}{NICE-SLAM}} & 
      \multirow{3}{*}{\rotatebox[origin=c]{90}{\makecell{Extra-\\polate}}}
          & PSNR $\uparrow$ 
          & \nd 23.83 & \nd 22.61 & \nd 21.97 & \nd 25.78 & \fs 25.30 & \nd 18.50 & \rd 22.82 & \fs 25.26 & \nd 23.26\\
          & & SSIM $\uparrow$ 
          & \nd 0.788 & \nd 0.813 & \nd 0.858 & \fs 0.887 & \fs 0.842 & \nd 0.826 & \nd 0.862 & \nd 0.875 & \nd 0.844 \\
          & & LPIPS $\downarrow$ 
          & \rd 0.284 & \nd 0.249 & \nd 0.218 & \nd 0.209 & \fs 0.145 & \nd 0.242 & \nd 0.190 & \fs 0.191 & \nd 0.216 \\
        \cdashlinelr{2-12}
      & \multirow{3}{*}{\rotatebox[origin=c]{90}{\makecell{Inter-\\polate}}}
          & PSNR $\uparrow$ 
          & \rd 22.12 & \rd 22.47 & \nd 24.52 & \fs 29.07 & \fs 30.34 & 19.66 & \rd 22.23 & \rd 24.94 & \nd 24.42
          \\
          & & SSIM $\uparrow$ 
          & 0.689 & \rd 0.757 & \rd 0.814 & 0.874 & \fs 0.886 & \rd 0.797 & 0.801 & \nd 0.856 & \rd 0.809
          \\
          & & LPIPS $\downarrow$ 
          & 0.330 & 0.271 & \nd 0.208 & \nd 0.229 & \nd 0.181 & \rd 0.235 & \nd 0.209 & \nd 0.198 & \nd 0.233
          \\
      \midrule
      \multirow{6}{*}{\rotatebox[origin=c]{90}{Vox-Fusion}} & 
      \multirow{3}{*}{\rotatebox[origin=c]{90}{\makecell{Extra-\\polate}}}
          & PSNR $\uparrow$ 
          & \rd 23.45 & 20.83 & \rd 18.38 & 23.28 & \nd 24.48 & \rd 17.50 & \nd 23.06 & \rd 24.84 & \rd 21.98 \\
          & & SSIM $\uparrow$ 
          & 0.765 & 0.773 & 0.747 & 0.751 & \rd 0.762 & 0.727 & \rd 0.824 & \rd 0.851 & \rd 0.775 \\
          & & LPIPS $\downarrow$ 
          & \nd 0.280 & \rd 0.272 & \rd 0.282 & \rd 0.235 & \rd 0.169 & \rd 0.292 & \rd 0.232 & \rd 0.212 & \rd 0.247 \\
        \cdashlinelr{2-12}
      & \multirow{3}{*}{\rotatebox[origin=c]{90}{\makecell{Inter-\\polate}}}
          & PSNR $\uparrow$ 
          & \nd 22.39 & \rd 22.36 & \rd 23.92 & \rd 27.79 & \nd 29.83 & \rd 20.33 & \nd 23.47 & \nd 25.21 & \rd 24.41
          \\
          & & SSIM $\uparrow$ 
          & 0.683 & 0.751 & 0.798 & \rd 0.857 & \nd 0.876 & 0.794 & \rd 0.803 & \rd 0.847 & 0.801
          \\
          & & LPIPS $\downarrow$ 
          & \nd 0.303 & \nd 0.269 & 0.234 & 0.241 & \rd 0.184 & 0.243 & \rd 0.213 & \rd 0.199 & \rd 0.236
          \\
      \midrule      
      \rowcolor{colorSep}
      \multicolumn{12}{l}{\textit{\textbf{RGB input}}}\\

\multirow{6}{*}{\rotatebox[origin=c]{90}{COLMAP}} & 
      \multirow{3}{*}{\rotatebox[origin=c]{90}{\makecell{Extra-\\polate}}}
          & PSNR $\uparrow$ 
          & 20.93 & 11.67 & 10.35 & 5.88 & 5.88 & 15.66 & 13.73 & 17.47 & 12.70  \\
          & & SSIM $\uparrow$ 
          & \rd 0.778 & 0.698 & 0.757 & 0.595 & 0.539 & \rd 0.806 & 0.792 & 0.811 & 0.722 \\
          & & LPIPS $\downarrow$ 
          & 0.291 & 0.443 & 0.330 & 0.444 & 0.337 & 0.303 & 0.299 & 0.273 & 0.340 \\
        \cdashlinelr{2-12}
      & \multirow{3}{*}{\rotatebox[origin=c]{90}{\makecell{Inter-\\polate}}}
          & PSNR $\uparrow$ 
          & 20.31 & 12.33 & 14.53 & 9.39 & 7.26 & 15.42 & 15.28 & 16.98 & 13.94 
          \\
          & & SSIM $\uparrow$ 
          & \nd 0.710 & 0.607 & 0.771 & 0.744 & 0.676 & 0.767 & 0.754 & 0.765 & 0.724
          \\
          & & LPIPS $\downarrow$ 
          & 0.332 & 0.531 & 0.318 & 0.377 & 0.306 & 0.341 & 0.330 & 0.290 & 0.353
          \\
      \midrule
      
      \multirow{6}{*}{\rotatebox[origin=c]{90}{DROID-SLAM}} & 
      \multirow{3}{*}{\rotatebox[origin=c]{90}{\makecell{Extra-\\polate}}}
          & PSNR $\uparrow$ 
          & 18.25 & 18.65 & 13.49 & 16.13 & 10.31 & 14.78 & 15.53 & 15.71 & 15.36 \\
          & & SSIM $\uparrow$ 
          & 0.737 & \rd 0.793 & \rd 0.786 & \rd 0.760 & 0.650 & 0.800 & 0.797 & 0.800 & 0.765 \\
          & & LPIPS $\downarrow$ 
          & 0.352 & 0.283 & 0.299 & 0.298 & 0.286 & 0.300 & 0.302 & 0.311 & 0.304 \\
        \cdashlinelr{2-12}
      & \multirow{3}{*}{\rotatebox[origin=c]{90}{\makecell{Inter-\\polate}}}
          & PSNR $\uparrow$ 
          & 21.41 & \fs 24.04 & 22.08 & \rd 23.59 & 21.29 & \nd 20.64 & 20.22 & 20.22 & 21.69
          \\
          & & SSIM $\uparrow$ 
          & \rd 0.693 & \fs 0.786 & \nd 0.826 & \fs 0.868 & \rd 0.863 & \fs 0.828 & \nd 0.808 & 0.819 & \nd 0.812
          \\
          & & LPIPS $\downarrow$ 
          & \rd 0.329 & \rd 0.270 & \rd 0.228 & \rd 0.232 & 0.207 & \nd 0.231 & 0.234 & 0.237 & 0.246
          \\
      \midrule
      \multirow{6}{*}{\rotatebox[origin=c]{90}{\textbf{NICER-SLAM}}} & 
      \multirow{3}{*}{\rotatebox[origin=c]{90}{\makecell{Extra-\\polate}}}
          & PSNR $\uparrow$ 
          &  \fs 25.64 & \fs 23.69 & \fs 22.62 & \fs 25.88 & \rd 22.56 & \fs 21.46 & \fs 24.42 & \nd 25.15 & \fs 23.93 \\
          & & SSIM $\uparrow$ 
          & \fs 0.810 & \fs 0.820 & \fs 0.871 & \nd 0.885 & \nd 0.828 & \fs 0.863 & \fs 0.888 & \fs 0.887 & \fs 0.857 \\
          & & LPIPS $\downarrow$ 
          & \fs 0.254 & \fs 0.233 & \fs 0.200 & \fs 0.193 & \nd 0.160 & \fs 0.203 & \fs 0.175 & \nd 0.192 & \fs 0.201 \\
        \cdashlinelr{2-12}
     & \multirow{3}{*}{\rotatebox[origin=c]{90}{\makecell{Inter-\\polate}}}
          & PSNR $\uparrow$ 
          & \fs 25.33 & \nd 23.92 & \fs 26.12 & \nd 28.54 & \rd 25.86 & \fs 21.95 & \fs 26.13 & \fs 25.47 & \fs 25.41 \\
          & & SSIM $\uparrow$ 
          & \fs 0.751 & \nd 0.771 & \fs 0.831 & \nd 0.866 & 0.852 & \nd 0.820 & \fs 0.856 & \fs 0.865 & \fs 0.827 \\
          & & LPIPS $\downarrow$ 
          & \fs 0.250 & \fs 0.215 & \fs 0.176 &  \fs 0.172 & \fs 0.178 & \fs 0.195 & \fs 0.162 & \fs 0.177 & \fs 0.191 \\
      \bottomrule
    \end{tabular}}%
    \caption{\textbf{Novel View Synthesis Evaluation on Replica dataset~\cite{replica19arxiv}.} Best results are highlighted as \colorbox{colorFst}{\bf first}, \colorbox{colorSnd}{second}, and \colorbox{colorTrd}{third}.
    Our method mostly outperforms all other methods even the ones that additionally use depth inputs.
    }
    \label{tab:replica_rendering}
    \vspace{-10pt}
\end{table}
}

\newcommand{\tablereplicareonstruction}{
\begin{table}[h]
  \centering
  \footnotesize
  \setlength{\tabcolsep}{1pt}
  \renewcommand{\arraystretch}{1.4}      %
  \resizebox{\linewidth}{!}{
  \begin{tabular}{llcccccccccccccccccc}
    \toprule
    & & \multicolumn{1}{c}{\makecell{\tt{rm-0}}} & \multicolumn{1}{c}{\makecell{\tt{rm-1}}} &  \multicolumn{1}{c}{\makecell{\tt{rm-2}}} & \multicolumn{1}{c}{\makecell{\tt{off-0}}} & \multicolumn{1}{c}{\makecell{\tt{off-1}}} & \multicolumn{1}{c}{\makecell{\tt{off-2}}}& \multicolumn{1}{c}{\makecell{\tt{off-3}}} & \multicolumn{1}{c}{\makecell{\tt{off-4}}} & Avg. \\
    \midrule    
    \rowcolor{colorSep}
    \multicolumn{12}{l}{\textit{\textbf{RGB-D input}}}\\
    \multirow{4}{*}{\rotatebox[origin=c]{90}{NICE-SLAM}}
    & Acc.[cm]$\downarrow$ & \nd 3.53 & \nd 3.60 & \fs 3.03 & 5.56 & \rd 3.35 & 4.71 & \nd 3.84 & \rd 3.35 & \rd 3.87 \\
    & Comp.[cm]$\downarrow$ & 3.40 & \nd 3.62 & \fs 3.27 & \fs 4.55 & \fs 4.03 & \fs 3.94 & \nd 3.99 & \rd 4.15 & \fs 3.87 \\
    & Comp.Ratio[$<\!5$cm\,\%]$\uparrow$ & \rd 86.05 & \nd 80.75 & \fs 87.23 & \nd 79.34 & \fs 82.13 & \rd 80.35 & \rd 80.55 & 82.88 & \nd 82.41 \\
    & Normal Cons.[\%]$\uparrow$ & \rd 91.92 & \rd 91.36 & \rd 90.79 & \nd 89.30 & \nd 88.79 & \rd 88.97 & \rd 87.18 & \nd 91.17 & \rd 89.93\\
    \midrule
     \multirow{4}{*}{\rotatebox[origin=c]{90}{Vox-Fusion}} 
    & Acc.[cm]$\downarrow$ & \fs 2.53 & \fs 1.69 & \rd 3.33 & \fs 2.20 & \fs 2.21 & \fs 2.72 & \rd 4.16 & \fs 2.48 & \fs 2.67 \\
    & Comp.[cm]$\downarrow$ & \fs 2.81 & \fs 2.51 & \rd 4.03 & \rd 8.75 & \rd 7.36 & \nd 4.19 & \fs 3.26 & \fs 3.49 & \rd 4.55 \\
    & Comp.Ratio[$<\!5$cm\,\%]$\uparrow$ & \fs 91.52 & \fs 91.34 & \nd 86.78 & \fs 81.99 & \nd 82.03 & \fs 85.45 & \fs 87.13 & \fs 86.53 & \fs 86.59 \\
    & Normal Cons.[\%]$\uparrow$ & \fs 94.14 & \fs 93.28 & \nd 91.71 & \fs 90.52 & \fs 88.95 & \fs 91.54 & \nd 91.03 & \fs 92.67 & \fs 91.73\\
    \midrule
    \rowcolor{colorSep}
    \multicolumn{12}{l}{\textit{\textbf{RGB input}}}\\
    \multirow{4}{*}{\rotatebox[origin=c]{90}{COLMAP}}
    & Acc.[cm]$\downarrow$ & \rd 3.87 & 27.29 & 5.41 & \rd 5.21 & 12.69 & \rd 4.28 & 5.29 & 5.45 & 8.69 \\
    & Comp.[cm]$\downarrow$ & 4.78 & 23.90 & 17.42 & 12.98 & 12.35 & 4.96 & 16.17 & 4.41 & 12.12 \\
    & Comp.Ratio[$<\!5$cm\,\%]$\uparrow$ & 83.08 & 22.89 & 64.47 & \rd 72.59 & 69.52 & \nd 81.12 & 64.38 & \rd 82.92 & 67.62 \\
    & Normal Cons.[\%]$\uparrow$ & 72.49 & 60.10 & 69.42 & 69.91 & 74.04 & 71.84 & 71.49 & 71.75 & 70.13 \\
    \midrule
    \multirow{4}{*}{\rotatebox[origin=c]{90}{DROID-SLAM}}
    & Acc.[cm]$\downarrow$ & 12.18 & 8.35 & \nd 3.26 & \nd 3.01 & \nd 2.39 & 5.66 & 4.49 & 4.65 & 5.50   \\
    & Comp.[cm]$\downarrow$ & 8.96 & 6.07 & 16.01 & 16.19 & 16.20 & 15.56 & 9.73 & 9.63 & 12.29 \\
    & Comp.Ratio[$<\!5$cm\,\%]$\uparrow$ & 60.07 & 76.20 & 61.62 & 64.19 & 60.63 & 56.78 & 61.95 & 67.51 & 63.62 \\
    & Normal Cons.[\%]$\uparrow$ & 72.81 & 74.71 & 79.21 & 77.53 & 78.57 & 75.79 & 77.69 & 76.38 & 76.59 \\
    \midrule    
    
    \multirow{4}{*}{\rotatebox[origin=c]{90}{\textbf{NICER-SLAM}}}
    & Acc.[cm]$\downarrow$ & \fs 2.53 & \rd 3.93 & 3.40 & 5.49 & 3.45 & \nd 4.02 & \fs 3.34 & \nd 3.03 & \nd 3.65 \\
    & Comp.[cm]$\downarrow$ & 3.04 & \rd 4.10 & \nd 3.42 & \nd 6.09 & \nd 4.42 & \rd 4.29 & \rd 4.03 & \nd 3.87 & \nd 4.16\\
    & Comp.Ratio[$<\!5$cm\,\%]$\uparrow$ & \nd 88.75 & \rd 76.61 & \rd 86.10 & 65.19 & \rd 77.84 & 74.51 & \nd 82.01 & \nd 83.98 & \rd 79.37\\
    & Normal Cons.[\%]$\uparrow$ & \nd 93.00 & \nd 91.52 & \fs 92.38 & \rd 87.11 & \rd 86.79 & \nd 90.19 & \nd 90.10 & \rd 90.96 & \nd 90.27\\
    \bottomrule
  \end{tabular}}%
  \caption{\textbf{Reconstruction Results on the Replica dataset.} Best results are highlighted as \colorbox{colorFst}{\bf first}, \colorbox{colorSnd}{second}, and \colorbox{colorTrd}{third}.
  NICER-SLAM performs the best among RGB SLAM methods, and is also on par with RGB-D methods.
  }\vspace{-1em}
  \label{tab:replica_rec}
\end{table}
}

\newcommand{\tableablation}{
\begin{table}[htbp]
    \centering
    \footnotesize
    \setlength{\tabcolsep}{2pt}          %
    \resizebox{\linewidth}{!}{
    \begin{tabular}{lccccc}
    \toprule
      {}&ATE RMSE$\downarrow$ & Acc. $\downarrow$ & Comp. $\downarrow$ & Comp. Ratio $\uparrow$ & Normal Cons. $\uparrow$\\
      \midrule
       w/o $\cL_\text{depth}\qquad$ & 4.48 & 4.74 & 6.18 & 71.10 & 89.23\\
       w/o $\cL_\text{normal}$ & 3.22 & 7.25 & 6.98 & 51.07 & 86.64\\
       w/o $\cL_\text{warp}$ & \rd 2.96 & \rd 3.76 & \rd 4.60 & \rd 74.42 & \fs 91.32\\
       w/o $\cL_\text{flow}$ &  \nd 2.30 & \nd 3.31 & \nd 4.31 & \nd 81.10 & \nd 91.00 \\
       \textbf{Ours} & \fs 2.01 & \fs 3.03 & \fs 3.87 & \fs 83.98 & \rd 90.96 \\
       \bottomrule
    \end{tabular}}\\
    (a) {\normalsize Ablation Study on Losses in~\eqnref{eq:loss_mapping}}.\vspace{0.7em}\\
    \resizebox{\linewidth}{!}{
    \begin{tabular}{lccccc}
    \toprule
      {}&ATE RMSE$\downarrow$ & Acc. $\downarrow$ & Comp. $\downarrow$ & Comp. Ratio $\uparrow$ & Normal Cons. $\uparrow$\\
      \midrule
       w/o $\{\Phi_l^\text{color}\}^L_1$ \;\;&  \rd 9.92 & \rd 8.32 & \rd 8.49 & \rd 50.13 & \rd 87.84 \\
       w/o $\Phi^\text{coarse}$ & \nd 3.07 & \nd 4.51 & \nd 5.11 & \nd 67.29 & \nd 90.17\\
       \textbf{Ours} & \fs 2.01 & \fs 3.03 & \fs 3.87 & \fs 83.98 & \fs 90.96 \\
       \bottomrule
    \end{tabular}}\\
    (b) {\normalsize Ablation Study on Hierarchical Architecture}.\vspace{0.7em}\\
    \resizebox{\linewidth}{!}{
    \begin{tabular}{lccccc}
    \toprule
      {}&ATE RMSE$\downarrow$ & Acc. $\downarrow$ & Comp. $\downarrow$ & Comp. Ratio $\uparrow$ & Normal Cons. $\uparrow$\\
      \midrule
    {Fixed $\beta$=0.01} \;& 3.81 & 7.77 & 8.28 & 39.48 & 87.52 \\ 
    {Fixed $\beta$=0.001} & 3.98 & \rd 3.48 & 5.05 & \rd 76.67 & 90.39 \\
    {Global optim. $\beta$} & \rd 2.62 & 3.64 & \rd 4.53 & 76.35 & \nd 90.88 \\ 
    {Voxel size $32^3$} & 3.00 & \nd 3.19 & \nd 4.35 & \nd 81.90 & \rd 90.65 \\
    {Voxel size $128^3$} & \nd 2.16 & 4.35 & 4.87 & 68.96 & 90.40 \\
       \textbf{Ours} & \fs 2.01 & \fs 3.03 & \fs 3.87 & \fs 83.98 & \fs 90.96 \\
       \bottomrule
    \end{tabular}}\\
    (c) {\normalsize Ablation Study on SDF-to-Density Transformation}.\vspace{0.7em}\\
    \caption{\textbf{Ablation Study.} On a single randomly selected Replica scene (\texttt{office-4}), we evaluate both camera tracking and reconstruction. Best results are highlighted as\colorbox{colorFst}{\bf first},\colorbox{colorSnd}{second}, and\colorbox{colorTrd}{third}.}
    \label{tab:ablation}
\end{table}
}

\begin{document}

\title{NICER-SLAM: Neural Implicit Scene Encoding for RGB SLAM}

\author{
Zihan Zhu$^{1*}$ \qquad Songyou Peng$^{1, 2}$\thanks{Equal contribution.}\qquad Viktor Larsson$^{3}$ \qquad 
Zhaopeng Cui$^{4}$  \qquad Martin R. Oswald$^{1,5}$\\Andreas Geiger$^{6}$ \qquad Marc Pollefeys$^{1,7}$\vspace{0.5em}\\
$^{1}$ETH Z\"urich\qquad
$^{2}$MPI for Intelligent Systems, T\"ubingen \qquad
$^{3}$Lund University \\
$^{4}$State Key Lab of CAD\&CG, Zhejiang University\qquad 
$^{5}$University of Amsterdam\qquad \\
$^{6}$University of Tübingen, Tübingen AI Center\qquad 
$^{7}$Microsoft
}

\twocolumn[{%
\renewcommand\twocolumn[1][]{#1}%
\maketitle
\vspace{-2.0em}
    \includegraphics[width=\textwidth]{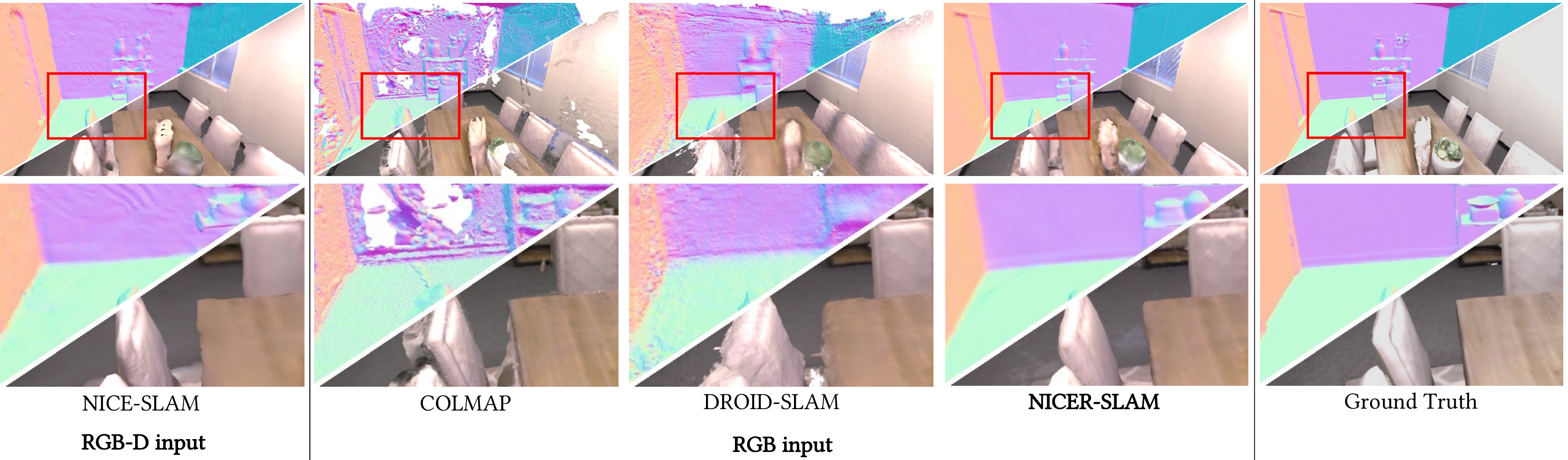}    
    \captionof{figure}{\small\textbf{3D Dense Reconstruction and Rendering from Different SLAM Systems.}
    On the Replica dataset~\cite{replica19arxiv}, we compare to dense RGB-D SLAM method NICE-SLAM~\cite{Zhu2022CVPR}, and monocular SLAM approaches COLMAP~\cite{Colmap2016}, DROID-SLAM~\cite{teed2021droid}, and our proposed NICER-SLAM.
  }
    \label{fig:teaser}
}]

\renewcommand*{\thefootnote}{\arabic{footnote}}

\ificcvfinal\thispagestyle{empty}\fi
\begin{abstract}
Neural implicit representations have recently become popular in simultaneous localization and mapping (SLAM), especially in dense visual SLAM.
However, previous works in this direction either rely on RGB-D sensors, or require a separate monocular SLAM approach for camera tracking and do not produce high-fidelity dense 3D scene reconstruction.
In this paper, we present NICER-SLAM, a dense RGB SLAM system that simultaneously optimizes for camera poses and a hierarchical neural implicit map representation, which also allows for high-quality novel view synthesis.
 To facilitate the optimization process for mapping, we integrate additional supervision signals including easy-to-obtain monocular geometric cues and optical flow, and also introduce a simple warping loss to further enforce geometry consistency.
 Moreover, to further boost performance in complicated indoor scenes, we also propose a local adaptive transformation from signed distance functions (SDFs) to density in the volume rendering equation.
 On both synthetic and real-world datasets we demonstrate strong performance in dense mapping, tracking, and novel view synthesis, even competitive with recent RGB-D SLAM systems.
\end{abstract}

\section{Introduction}\label{sec:intro}

Simultaneous localization and mapping (SLAM) is a fundamental computer vision problem with wide applications in autonomous driving, robotics, mixed reality, and more.
Many dense visual SLAM methods have been introduced in the past years~\cite{Newcombe2011KinectFusion,Schops2019CVPR,Whelan2012Kintinuous,Whelan2015RSS,Newcombe2011DTAM} and they are able to produce dense reconstructions of indoor scenes in real-time.
However, most of these approaches rely on RGB-D sensors and fail on outdoor scenes or when depth sensors are not available.
Moreover, for unobserved regions, they have difficulty making plausible geometry estimations.
In the deep learning era, a handful of dense monocular SLAM systems~\cite{Bloesch2018CVPR,Czarnowski2020RAL,Zhi2019CVPR} %
take only RGB sequences as input
and to some extent fill in unobserved regions due to their monocular depth prediction networks.
Nevertheless, these systems are typically only applicable to small scenes with limited camera movements.

With the rapid developments in neural implicit representations or neural fields~\cite{Xie2022CGF}, they have also demonstrated powerful performance in end-to-end differentiable dense visual SLAM.
iMAP~\cite{Sucar2021ICCV} first shows the capability of neural implicit representations in dense RGB-D SLAM, but it is only limited to room-size datasets.
NICE-SLAM~\cite{Zhu2022CVPR} introduces a hierarchical implicit encoding to perform mapping and camera tracking in much larger indoor scenes. 
Although follow-up works~\cite{Yang2022ISMAR,Ming2022ARXIV,Kruzhkov2022MESLAM,Johari2022ESLAM,Lisus2023ARXIV,Ortiz2022RSS} try to improve upon NICE-SLAM and iMAP from different perspectives, all of these works still rely on the reliable depth input from RGB-D sensors.

Very recently, a handful of concurrent works (available as pre-prints) try to apply neural implicit representations for RGB-only SLAM~\cite{Rosinol2022ARXIV, Chung2022ARXIV}.
However, their tracking and mapping pipelines are independent of each other as they rely on different scene representations for these tasks. 
Both approaches depend on the state-of-the-art visual odometry
methods~\cite{teed2021droid,Mur2017TRO} for camera tracking, while using neural radiance fields (NeRFs) only for mapping. 
Moreover, they both only output and evaluate the rendered depth maps and color images, so no dense 3D model of a scene is produced.
This raises an interesting research question:
\begin{center}
\emph{Can we build a unified dense SLAM system with a neural implicit scene representation for both tracking and mapping from a monocular RGB video?}
\end{center}
Compared to RGB-D SLAM, RGB-only SLAM is more challenging for multiple reasons.
\textbf{1)} Depth ambiguity: often several possible correspondences match the color observations well, especially with little available texture information.
Hence, stronger geometric priors are required for both mapping and tracking optimizations.
\textbf{2)} Harder 3D reconstruction: due to the ambiguity the estimation of surfaces is less localized, leading to more complex data structure updates and increased sampling efforts.
\textbf{3)} Optimization convergence: as a result of the previous challenges, the resulting optimization is less constrained and more complex - leading to slower convergence.

To tackle these challenges, we introduce \emph{NICER-SLAM}, an implicit-based dense RGB SLAM system that is end-to-end optimizable for both mapping and tracking, and also capable of learning an accurate scene representation for novel view synthesis.
Our key ideas are as follows. 
First, for scene geometry and colors, we present coarse-to-fine hierarchical feature grids to model the signed distance functions (SDFs), which yields detailed 3D reconstructions and high-fidelity renderings.
Second, to optimize neural implicit map representations with only RGB input, we integrate additional signals for supervision including easy-to-obtain monocular geometric cues and optical flow, and also introduce a simple warping loss to further enforce geometry consistency. 
We observe that those regularizations help to significantly disambiguate the optimization process, enabling our framework to work accurately and robustly with only RGB input.
Third, to better fit the sequential input for indoor scenes, we propose to use a locally adaptive transformation from SDF to density.

In summary, we make the following contributions:

\begin{itemize}[leftmargin=*]
    \item We present NICER-SLAM, one of the first dense RGB-only SLAM that is end-to-end optimizable for both tracking and mapping, which also allows for high-quality novel view synthesis.
    \item We introduce a hierarchical neural implicit encoding for SDF representations, different geometric and motion regularizations, as well as a locally adaptive SDF to volume density transformation. 
    \item We demonstrate strong performances in mapping, tracking, and novel view synthesis on both synthetic and real-world datasets, even competitive with recent RGB-D SLAM methods.
\end{itemize}

\section{Related Work}\label{sec:related}
\figpipeline

\paragraph{Dense Visual SLAM}
SLAM is an active field in both industry and academia, especially in the past two decades. 
While sparse visual SLAM algorithms~\cite{Mur2015TRO,Mur2017TRO,Engel2017PAMI,Klein2007ISMAR} estimate accurate camera poses and only have sparse point clouds as the map representation, dense visual SLAM approaches focus on recovering a dense map of a scene.
In general, map representations are categorized as either view-centric or world-centric.
The first often represents 3D geometry as depth maps for keyframes, including the seminal work DTAM~\cite{Newcombe2011DTAM}, as well as many follow-ups~\cite{Ummenhofer2017CVPR,Zhou2018ECCV,Teed2020ICLR,Czarnowski2020RAL,Tang2019ICLR,Bloesch2018CVPR,Zhi2019CVPR,Sucar2020THREEDV,teed2021droid,Koestler2022CORL}.
Another line of research considers world-centric maps, and they anchor the 3D geometry of a full scene in uniform world coordinates and represent as surfels~\cite{Whelan2015RSS,Schops2019CVPR} or occupancies/TSDF values in the voxel grids~\cite{Bylow2013RSS,Dai2017TOG,Niessner2013TOG,Newcombe2011KinectFusion}.
Our work also falls into this category and uses a world-centric map representation, but instead of explicitly representing the surface, we store latent codes in multi-resolution voxel grids.
This allows us to not only reconstruct high-quality geometry at low grid resolutions, but also attain plausible geometry estimation for unobserved regions.

\paragraph{Neural Implicit-based SLAM}
Neural implicit representations~\cite{Xie2022CGF} have shown great performance in many different tasks, including object-level reconstruction reconstruction~\cite{Mescheder2019CVPR,Chen2019CVPR,Park2019CVPR,Peng2021NEURIPS,Liu2020CVPR,Niemeyer20cvpr_DVR,yariv2020multiview}, scene completion~\cite{Peng2020ECCV,Lionar2021WACV,Jiang2020CVPR}, novel view synthesis~\cite{Mildenhall2020ECCV,Reiser2021ICCV,Zhang20arxiv_nerf++,Mueller2022SIGGRAPH,wu2022scalable}, etc.
In terms of SLAM-related applications, some works~\cite{yen2020inerf,lin2021barf,Wang2021ARXIV,Chng2022ECCV,Bian2022AXRIV,Clark2022CVPR} try to jointly optimize a neural radiance field and camera poses, but they are only applicable to small objects or small camera movements.
A series of recent works~\cite{Chung2022ARXIV,Rosinol2022ARXIV} relax such constraints, but they mainly rely on state-of-the-art SLAM systems like ORB-SLAM and DROID-SLAM to obtain accurate camera poses, and do not produce 3D dense reconstruction but only novel view synthesis results. 

iMAP~\cite{Sucar2021ICCV} and NICE-SLAM~\cite{Zhu2022CVPR} are the first two unified SLAM pipelines using neural implicit representations for both mapping and camera tracking. 
iMAP uses a single MLP as the scene representation so they are limited to small scenes, while NICE-SLAM can scale up to much larger indoor environments by applying hierarchical feature grids and tiny MLPs as the scene representation. 
Many follow-up works improve upon these two works from various perspectives, including efficient scene representation~\cite{Johari2022ESLAM,Kruzhkov2022MESLAM}, fast optimziation~\cite{Yang2022ISMAR}, add IMU measurements~\cite{Lisus2023ARXIV}, or different shape representations~\cite{Ortiz2022RSS,Ming2022ARXIV}. 
However, all of them require RGB-D inputs, which limits their applications in outdoor scenes or when only RGB sensors are available.
In comparison, given only RGB sequences as input, our proposed neural-implicit-based system can output high-quality 3D reconstruction and recover accurate camera poses simultaneously.

A concurrent work~\cite{Li2023ICLR}\footnote{Arxiv version published on Jan 24, 2023.} presents a system in a similar spirit to ours.
While they optimize for accurate camera tracking, our method focuses on high-quality 3D reconstruction and novel view synthesis.
\section{Method}\label{sec:method}

We provide an overview of the NICER-SLAM pipeline in~\figref{fig:system_overview}. 
Given an RGB video as input, we simultaneously estimate accurate 3D scene geometry and colors, as well as camera tracking via end-to-end optimization.
We represent the scene geometry and appearance using hierarchical neural implicit representations (\secref{sec:method_representation}).
With NeRF-like differentiable volume rendering, we can render color, depth, and normal values of every pixel (\secref{sec:method_render}), which will be used for end-to-end joint optimization for camera pose, scene geometry, and color (\secref{sec:method_optimization}).
Finally we discuss some of the design choices made in our system (\secref{sec:method_design}).

\subsection{Hierarchical Neural Implicit Representations}\label{sec:method_representation}
We first introduce our optimizable hierarchical scene representations that combine multi-level grid features with MLP decoders for SDF and color predictions.

\paragraph{Coarse-level Geometric Representation}
The goal of the coarse-level geometric representation is to efficiently model the coarse scene geometry (objects without capturing geometric details) and the scene layout (e.g. walls, floors) even with only partial observations.
To this end, we represent the normalized scene with a dense voxel grid with a resolution of $32\times32\times32$, and keep a 32-dim feature in each voxel. 
For any point $\bx\in\nR^3$ in the space, we use a small MLP $f^\text{coarse}$ with a single 64-dim hidden layer to obtain its base SDF value $s^\text{coarse} \in \nR$ and a geometric feature $\bz^\text{coarse}\in\nR^{32}$ as:
\begin{equation}
    {s^\text{coarse},\;\bz^\text{coarse}} = f^\text{coarse}\big(\gamma(\bx), \Phi^\text{coarse}(\bx)\big),
    \label{eq:base_sdf}
\end{equation}
where $\gamma$ corresponds to a fixed positional encoding~\cite{Mildenhall2020ECCV,Tancik2020NEURIPS} mapping the coordinate to higher dimension.
Following~\cite{Yu2022MonoSDF,yariv2020multiview,volsdf}, we set the level for positional encoding to 6. 
$\Phi^\text{coarse}(\bx)$ denotes that the feature grid $\Phi^\text{coarse}$ is tri-linearly interpolated at the point $\bx$.

\paragraph{Fine-level Geometric Representation}
While the coarse geometry can be obtained by our coarse-level shape representation, it is important to capture high-frequency geometric details in a scene.
To realize it, we further model the high-frequency geometric details as residual SDF values with multi-resolution feature grids and an MLP decoder~\cite{Chibane2020CVPR,Mueller2022SIGGRAPH,Zhu2022CVPR,Takikawa2021CVPR}.
Specifically, we employ multi-resolution dense feature grids $\{\Phi_l^\text{fine}\}^L_1$ with resolutions $R_l$.
The resolutions are sampled in geometric space~\cite{Mueller2022SIGGRAPH} to combine features at different frequencies: %
\begin{equation}
    R_l := \;\lfloor R_\text{min} b^l \rfloor \hspace{0.3cm}
    b := \;\text{exp}\left(\frac{\ln R_\text{max} - \ln R_\text{min}}{L-1}\right)
    \enspace,
    \label{eq:res}
\end{equation}
where $R_\text{min}, R_\text{max}$ correspond to the lowest and highest resolution, respectively.
Here we consider $R_\text{min}=32$, $R_\text{max}=128$, and in total $L=8$ levels.
The feature dimension is $4$ for each level.

Now, to model the residual SDF values for a point $\bx$, we extract and concatenate the tri-linearly interpolated features at each level, and input them to an MLP $f^\text{fine}$ with 3 hidden layers of size 64:
\begin{equation}
	(\, \Delta s,\;\bz^\text{fine} \,) = f^\text{fine}\big(\gamma(\bx),\; \{\Phi_l^\text{fine}(\bx)\}\big)
    \enspace,
	\label{eq:f_full}
\end{equation}
where $\bz^\text{fine}\in\nR^{32}$ is the geometric feature for $\bx$ at the fine level.

With the coarse-level base SDF value $s^\text{coarse}$ and the fine-level residual SDF $\Delta s$, the final predicted SDF value $\hat{s}$ for $\bx$ is simply the sum between two:
\begin{equation}
    \hat{s} = s^\text{coarse} + \Delta s  \enspace.
    \label{eq:fine_sdf}
\end{equation}

\paragraph{Color Representation} 
Besides 3D geometry, we also predict color values such that our mapping and camera tracking can be optimized also with color losses.
Moreover, as an additional application, we can also render images from novel views on the fly.
Inspired by~\cite{Mueller2022SIGGRAPH}, we encode the color using another multi-resolution feature grid $\{\Phi_l^\text{color}\}_1^L$ and a decoder $f^\text{color}$ parameterized with a 2-layer MLP of size 64.
The number of feature grid levels is now $L=16$, the features dimension is 2 at each level. 
The minimum and maximum resolution now become $R_\text{min}=16$ and $R_\text{max}=2048$, respectively.
We predict per-point color values as: 
\begin{equation}
    \hat{\bc} = f^\text{color}\big(\bx, \hat{\bn}, \gamma(\bv), \bz^\text{coarse}, \bz^\text{fine}, \{\Phi_l^\text{color}(\bx)\}\big) \enspace.
\end{equation}
where $\hat{\bn}$ corresponds to the normal at point $\bx$ calculated from $\hat{s}$ in~\eqnref{eq:fine_sdf}, and $\gamma(\bv)$ is the viewing direction with positional encoding with a level of 4, following~\cite{volsdf,Yu2022MonoSDF}.

\subsection{Volume Rendering}\label{sec:method_render}
Following recent works on implicit-based 3D reconstruction~\cite{Oechsle2021ICCV,volsdf,Yu2022MonoSDF,wang2021neus} and dense visual SLAM~\cite{Sucar2021ICCV,Zhu2022CVPR}, we optimize our scene representation from~\secref{sec:method_representation} using differentiable volume rendering.
More specifically, in order to render a pixel, we cast a ray $\br$ from the camera center $\bo$ through the pixel along its normalized view direction $\bv$. 
$N$ points are then sampled along the ray, denoted as $\bx_i = \bo + t_i\bv$, and their predicted SDFs and color values are $\hat{s}_i$ and $\hat{\bc}_i$, respectively. 
For volume rendering, we follow~\cite{volsdf} to transform the SDFs $\hat{s}_i$ to density values $\sigma_i$: 
\begin{equation}\label{eq:laplace}
  \sigma_\beta(s) = 
  \begin{cases} 
     \frac{1}{2\beta} \exp\big( \frac{s}{\beta} \big) & \text{if } s\leq 0 \\
     \frac{1}{\beta}\left( 1-\frac{1}{2}\exp\big( -\frac{s}{\beta} \big) \right) & \text{if } s>0
     \enspace,
  \end{cases}
\end{equation}
where $\beta \in \nR$ is a parameter controlling the transformation from SDF to volume density.
As in~\cite{Mildenhall2020ECCV}, the color $\hat{C}$ for the current ray $\br$ is calculated as:
\begin{equation}
\begin{gathered}
  \hat{C}  = \sum_{i=1}^N \, T_i \, \alpha_i \, \hat{\bc}_i 
  \hspace{0.5cm}
    T_i = \prod_{j=1}^{i-1}\left(1-\alpha_j\right) \\
\alpha_i = 1-\exp\left(-\sigma_i\delta_i\right) \enspace,
  \label{eq:volume_render}
\end{gathered}
\end{equation}
where $T_i$ and $\alpha_i$ correspond to transmittance and alpha value of sample point $i$ along ray $\br$, respectively,
and $\delta_i$ is the distance between neighboring sample points. 
In a similar manner, we can also compute the depth $\hat{D}$ and normal $\hat{N}$ of the surface intersecting the current ray $\br$ as:
\begin{equation}
  \hat{D} = \sum_{i=1}^N \, T_i \, \alpha_i \, t_i \hspace{1cm}
  \hat{N} = \sum_{i=1}^N \, T_i \, \alpha_i \, \hat{\bn}_i  \enspace.
  \label{eq:volume_render_dn}
\end{equation}

\paragraph{Locally Adaptive Transformation}
The $\beta$ parameter in \eqnref{eq:laplace} models the smoothing amount near the object’s surface.
The value of $\beta$ gradually decreases during the optimization process as the network becomes more certain about the object surface. 
Therefore, this optimization scheme results in faster and sharper reconstructions.

In VolSDF~\cite{volsdf}, they model $\beta$ as a single global parameter.
This way of modeling the transformation intrinsically assumes that the degree of optimization is the same across different regions of the scene, which is sufficient for small object-level scenes.
However, for our sequential input setting within a complicated indoor scene, a global optimizable $\beta$ is sub-optimal (\cf ablation study in~\secref{sec:exp_ablation}). 
Therefore, we propose to assign $\beta$ values \emph{locally} so the SDF-density transformation in~\eqnref{eq:laplace} is also locally adaptive.

In detail, we maintain a voxel counter across the scene and count the number of point samples within every voxel during the mapping process. We empirically choose the voxel size of $64^3$ (see ablation study in \secref{sec:exp_ablation}).
Next, we heuristically design a transformation from local point samples counts $T_{p}$ to the $\beta$ value:
\begin{equation}
  \beta = c_0\cdot\exp( -c_1\cdot T_{p})+c_2  \enspace.
  \label{eq:adaptive_transform}
\end{equation}

We come up with the transformation by plotting $\beta$ decreasing curve with respect to the voxel count under the global input setting, and fitting a function on the curve. We empirically find that the exponential curve is the best fit. 

\subsection{End-to-End Joint Mapping and Tracking}\label{sec:method_optimization}
Purely from an RGB sequential input, it is very difficult to jointly optimize 3D scene geometry and color together with camera poses, due to the high degree of ambiguity, especially for large complex scenes with many textureless and sparsely covered regions.
Therefore, to enable end-to-end joint mapping and tracking under our neural scene representation, we propose to use the following loss functions, including geometric and prior constraints, single- and multi-view constraints, as well as both global and local constraints.

\paragraph{RGB Rendering Loss}
\eqnref{eq:volume_render} connects the 3D neural scene representation to 2D observations, hence, we can optimize the scene representation with a simple RGB reconstruction loss:
\begin{equation}
  \mathcal{L}_\text{rgb} = \sum_{\br \in \cR} {\lVert \hat{C}(\br) - C(\br) \rVert}_1 \enspace ,
  \label{eq:loss_rgb}
\end{equation}
where $\cR$ denotes the randomly sampled pixels/rays in every iteration, and $C$ is the input pixel color value.

\paragraph{RGB Warping Loss}
To further enforce geometry consistency from only color inputs, we also add a simple per-pixel warping loss.

For a pixel in frame $m$, denoted as $\br_m$, we first render its depth value using~\eqnref{eq:volume_render_dn} and unproject it to 3D, and then project it to another frame $n$ using intrinsic and extrinsic parameters of frame $n$.
The projected pixel $\br_{m\rightarrow n}$ in the nearby keyframe $n$ is denoted as $\br_{m\rightarrow n}$.
The warping loss is then defined as:
\begin{equation}
  \mathcal{L}_\text{warp} = \sum_{\br_m \in \cR}\sum_{n\in\cK_m} {\lVert C(\br_m) - C(\br_{m\rightarrow n}) \rVert}_1 \enspace ,
\end{equation}
where $\cK_m$ denotes the keyframe list for the current frame $m$, excluding frame $m$ itself.
We mask out the pixels that are projected outside the image boundary of frame $n$.
Note that unlike~\cite{Darmon2022CVPR} that optimize neural implicit surfaces with patch warping, we observed that it is way more efficient and without performance drop to simply perform warping on randomly sampled pixels.

\paragraph{Optical Flow Loss}
Both the RGB rendering and the warping loss are only point-wise terms which are prone to local minima. 
We therefore add a loss that is based on optical flow estimates which adhere to regional smoothness priors and help to tackle ambiguities.
Suppose the sample pixel in frame $m$ as $\br_m$ and the corresponding projected pixel in frame $n$ as $\br_n$, we can add an optical flow loss as follows:
\begin{equation}
  \mathcal{L}_\text{flow} = \sum_{\br_m \in \cR}\sum_{n\in\cK_m} {\lVert (\br_m - \br_n) - \text{GM}(\br_{m\rightarrow n}) \rVert}_1 \enspace ,
\end{equation}
where $\text{GM}(\br_{m\rightarrow n})$ denotes the estimated optical flow from GMFlow~\cite{xu2022gmflow}.

\paragraph{Monocular Depth Loss}
Given RGB input, one can easily obtain geometric cues (such as depths or normals) via an off-the-shelf monocular predictor~\cite{Eftekhar2021ICCV}.
Inspired by~\cite{Yu2022MonoSDF}, we also include this information into the optimization to guide the neural implicit surface reconstruction.

More specifically, to enforce depth consistency between our rendered expected depths $\hat{D}$ and the monocular depths $\bar{D}$,  we use the following loss~\cite{Ranftl2020PAMI,Yu2022MonoSDF}:
\begin{equation}
  \mathcal{L}_\text{depth} = \sum_{\br \in \cR} {\lVert (w \hat{D}(\br) + q) - \bar{D}(\br) \rVert}^2
  \label{eq:depth_prior} \enspace,
\end{equation}
where $w, q \in \nR$ are the scale and shift used to align $\hat{D}$ and $\bar{D}$, since $\bar{D}$ is only known up to an unknown scale. 
We solve for $w$ and $q$ per image with a least-squares criterion~\cite{Ranftl2020PAMI}, which has a closed-form solution.

\paragraph{Monocular Normal Loss}
Another geometric cue that is complementary to the monocular depth is surface normal.
Unlike monocular depths that provide global surface information for the current view, surface normals are local and capture more geometric details.
Similar to~\cite{Yu2022MonoSDF}, we impose consistency on the volume-rendered normal $\hat{N}$ and the monocular normals $\bar{N}$ from~\cite{Eftekhar2021ICCV} with angular and L1 losses:

\begin{equation}
\begin{aligned}
  \mathcal{L}_\text{normal} = \sum_{\br \in \cR} &{\lVert \hat{N}(\br) - \bar{N}(\br) \rVert}_1 \\
   + &  {\lVert  1 - \hat{N}(\br)^\top  \bar{N}(\br) \rVert}_1 \enspace .
\end{aligned}
\end{equation}

\paragraph{Eikonal Loss}
In addition, we add the Eikonal loss~\cite{gropp2020implicit} to regularize the output SDF values $\hat{s}$:
\begin{equation}
\mathcal{L}_\text{eikonal} = \sum_{\bx \in \cX} ({\lVert \nabla \hat{s}(\bx) \rVert}_2 - 1)^2 \enspace ,
\end{equation} 
where $\cX$ are a set of uniformly sampled near-surface points.

\paragraph{Optimization Scheme}
Finally, we provide details on how to optimize the scene geometry and appearance in the form of our hierarchical representation, and also the camera poses.

\paragraphcolon{\textit{Mapping}}
To optimize the scene representation mentioned in~\secref{sec:method_representation}, we uniformly sample $M$ pixels/rays in total from the current frame and selected keyframes. 
Next, we perform a 3-stage optimization similar to~\cite{Zhu2022CVPR} but use the following loss:
\begin{equation}
\begin{split}
   \cL \!=\! \cL_\text{rgb}&+0.5\cL_\text{warp}+0.001\cL_\text{flow}\\
                &+0.1\cL_\text{depth}+0.05\cL_\text{normal}+0.1\cL_\text{eikonal}
  \label{eq:loss_mapping}
\end{split}
\end{equation}

At the first stage, we treat the coarse-level base SDF value $s^\text{coarse}$ in~\eqnref{eq:base_sdf} as the final SDF value $\hat{s}$, and optimize the coarse feature grid $\Phi^\text{coarse}$, coarse MLP parameters of $f^\text{coarse}$, and color MLP parameters of $f^\text{color}$ with~\eqnref{eq:loss_mapping}.
Next, after $25\%$ of the total number of iterations, we start using~\eqnref{eq:fine_sdf} as the final SDF value so the fine-level feature grids $\{\Phi_l^\text{fine}\}$ and fine-level MLP $f^\text{fine}$ are also jointly optimized. 
Finally, after $75\%$ of the total number of iterations, we conduct a local bundle adjustment (BA) with~\eqnref{eq:loss_mapping}, where we also include the optimization of color feature grids $\{\Phi_l^\text{color}\}$ as well as the extrinsic parameters of $K$ selected mapping frames.

\paragraphcolon{\textit{Camera Tracking}}
We run in parallel camera tracking to optimize the camera pose (rotation and translation) of the current frame, while keeping the hierarchical scene representation fixed. 
Straightforwardly, we sample $M_{t}$ pixels from the current frame and use purely the RGB rendering loss in~\eqnref{eq:loss_rgb} for 100 iterations.

\subsection{System Design} \label{sec:method_design}

\paragraph{Frame Selection for Mapping}
During the mapping process, we need to select multiple frames from which we sample rays and pixels.
We introduce a simple frame selection mechanism.
First, we maintain a global keyframe list for which we directly add one every 10 frames.
For mapping, we select in total $K=16$ frames, where 5 of them are randomly selected from the keyframe list, 10 are randomly selected from the latest 20 keyframes, and also the current frame.

\paragraph{Implementation Details}
Mapping is done every 5 frames, while tracking is done every frame.
Note that with the pure RGB input, drifting is a known hard problem. 
To alleviate this, during the local BA stage in mapping, we freeze the camera poses for half of the 16 selected frames which are far from the current frame, and only optimize the camera poses of the other half jointly with the scene representation.
For the adaptive local transformation in \eqnref{eq:adaptive_transform}, we set $c_0=1.208\cdot 10^{-2}$, $c_1=6.26471\cdot 10^{-6}$ and $c_2=2.3\cdot10^{-3}$. 
For mapping and tracking, we sample $M=8096$ and $M_t=1024$ pixels, respectively, and optimize for 100 iterations. 
For every mapping iteration, we randomly sample pixels from all selected frames, instead of one frame per iteration. 
With our current unoptimized PyTorch implementation, it takes on average 496ms and 147ms for each mapping and tracking iteration on a single A100 GPU.
The coarse geometry is initialized as a sphere following~\cite{Yu2022MonoSDF, volsdf, yariv2020multiview}. To optimize the scene representation during the first mapping step, we assign the scale and shift value as $w=20$ and $q=0$ in \eqnref{eq:depth_prior}, so that the scaled monocular depth values are roughly reasonable (e.g. 1-5 meters).
The final mesh is extracted from the scene representation using Marching Cubes~\cite{lorensen1987marching} with a resolution of $512^3$.

\section{Experiments}\label{sec:experiments}

We evaluate qualitative and quantitative comparisons against state-of-the-art (SOTA) SLAM frameworks on both synthetic and real-world datasets in~\secref{sec:exp_evaluation}. 
A comprehensive ablation study that supports our design choices is also provided in~\secref{sec:exp_ablation}.

\paragraph{Datasets} 
We first evaluate on a synthetic dataset Replica~\cite{replica19arxiv}, where RGB-(D) images can be rendered out with the official renderer. 
To verify the robustness of our approach, we compare a challenging real-world dataset 7-Scenes~\cite{7scenes}. 
It consists of low-resolution images with severe motion blurs.
We use COLMAP to obtain the intrinsic parameters of the RGB camera.

\paragraph{Baselines}
We compare to 
(a) SOTA neural implicit-based RGB-D SLAM system NICE-SLAM~\cite{Zhu2022CVPR} and Vox-Fusion \cite{Yang2022ISMAR}, 
(b) classic MVS method COLMAP~\cite{Colmap2016}, and 
(c) SOTA dense monocular SLAM system DROID-SLAM~\cite{teed2021droid}. 
For camera tracking evaluation, we also compare to DROID-SLAM$^*$, which does not perform the final global bundle adjustment and loop closure (identical to our NICER-SLAM setting). 
For DROID-SLAM's 3D reconstruction, we run TSDF fusion with their predicted depths of keyframes.

\paragraph{Metrics} 
For camera tracking, we follow the conventional monocular SLAM evaluation pipeline where the estimated trajectory is aligned to the GT trajectory using \texttt{evo}~\cite{grupp2017evo}, and then
evaluate the accuracy of camera tracking (\textit{ATE RMSE})~\cite{sturm2012benchmark}.
To evaluate scene geometry, we consider \textit{Accuracy}, \textit{Completion}, \textit{Completion Ratio}, and \textit{Normal Consistency}.
The reconstructed meshes from monocular SLAM systems are aligned to the GT mesh using the ICP tool from~\cite{girardeau2016cloudcompare}. 
Moreover, we use PSNR, SSIM~\cite{ssim} and LPIPS~\cite{lpips} for novel view synthesis evaluation.

\tablereplicareonstruction
\figreplicareconstruction
\tablereplicatracking
\tablereplicarendering

\subsection{Mapping, Tracking and Rendering Evaluations}\label{sec:exp_evaluation}

\paragraph{Evaluation on Replica~\cite{replica19arxiv}}
First of all, for the evaluation of scene geometry, as shown in~\tabref{tab:replica_rec}, our method significantly outperforms RGB-only baselines like DROID-SLAM and COLMAP, and even shows competitive performance against RGB-D SLAM approaches like NICE-SLAM and Vox-Fusion.
Moreover, due to the use of monocular depth and normal cues, we show in~\figref{fig:replica_rec} that our reconstructions can faithfully recover geometric details and attain the most visually appealing results among all methods.
For camera tracking, we can clearly see on~\tabref{tab:replica_tracking} that the state-of-the-art method DROID-SLAM outperforms all methods including those RGB-D SLAM systems.
Nevertheless, our method is still on par with NICE-SLAM (1.88 vs 1.95 cm on average), while no ground truth depths are used as additional input.
\figreplicarendering
\figsevenscenesreconstruction

It is worth noting that even with the less accurate camera poses from our tracking pipeline, our results in novel view synthesis are notably better than all baseline methods including those using additional depth inputs, see~\tabref{tab:replica_rendering} and~\figref{fig:replica_rendering}.
On the one hand, the methods using traditional representations like COLMAP and DROID-SLAM suffer from rendering missing regions in the 3D reconstruction, and their renderings tend to be noisy.
On the other hand, neural-implicit based approaches like NICE-SLAM and Vox-Fusion can fill in those missing areas but their renderings are normally over-smooth.
We can faithfully render high-fidelity novel views even when those views are far from the training views.
This illustrates the effectiveness of different losses for disambiguating the optimization of our scene representations.

\tablesevenscenestracking
\paragraph{Evaluation on 7-Scenes~\cite{7scenes}}
We also evaluate on the challenging real-world dataset 7-Scenes to benchmark the robustness of different methods when the input images are of low resolutions and have severe motion blurs.
For geometry illustrated in~\figref{fig:7scenes}, we can notice that NICER-SLAM produces sharper and more detailed geometry over all baselines.
In terms of tracking, as can be observed in~\tabref{tab:7scenes_tracking}, baselines with RGB-D input outperform RGB-only methods overall, indicating that the additional depth inputs play an essential role for tracking especially when RGB images are imperfect.
Among RGB-only methods, COLMAP and DROID-SLAM* (without global bundle adjustment) perform poorly in the \texttt{pumpkin} scene because it contains large textureless and reflective regions in the RGB sequence.
NICER-SLAM is more robust to such issues thanks to the predicted monocular geometric priors.

\subsection{Ablation Study}\label{sec:exp_ablation}
To support our design choices, we investigate the effectiveness of different losses, the hierarchical architecture, SDF-density transformation, as well as the comparison between SDF and occupancy.

\paragraph{Losses}
We first verify the effectiveness of different losses for the mapping process from~\secref{sec:method_optimization}.
In \tabref{tab:ablation}~(a), we evaluate both 3D reconstruction and tracking because we conduct local BA on the third stage of mapping.
As can be noticed, using all losses together leads to the best overall performance. 
Without monocular depth or normal loss, both mapping and tracking accuracy drops significantly, indicating that these monocular geometric cues are important for the disambiguation of the optimization process.

\paragraph{Hierarchical Architecture}
In \tabref{tab:ablation}~(b) we compare our proposed scene representations to two variations. 
The first one is to remove the multi-resolution color feature grids $\{\Phi_l^\text{color}\}_1^L$ and only represent scene colors with the MLP $f^\text{color}$.
This change leads to large performance drops on all metrics, showing the necessity of having multi-res feature grids for colors.
The second variation is to remove the coarse feature grid $\Phi^\text{coarse}$ and uses only fine-level feature grids to represent SDFs. 
This also causes inferior performance, especially in the completeness/completeness ratio, indicating that the coarse feature grid can indeed help to learn geometry better.

\paragraph{SDF-to-Density Transformation}
We also compare different design choices for the transformation from SDF to volume density (see \secref{sec:method_render}):
(a) Fixed $\beta$ value, (b) globally optimizable $\beta$ as in~\cite{volsdf}, and also (c) different voxel size for counting (our default setting uses $64^3$).
As can be seen in~\tabref{tab:ablation}~(c), with the locally adaptive transformation and under the chosen voxel size, our method is able to obtain both better scene geometry and camera tracking.

\paragraph{SDF vs. Occupancy}
Unlike recent implicit-based dense SLAM systems~\cite{Sucar2021ICCV,Zhu2022CVPR,Yang2022ISMAR} which use occupancy to implicitly represent scene geometry, we instead use SDFs.
To verify this design choice, we keep the architecture identical but only replace the output in~\eqnref{eq:fine_sdf} to let the occupancy probability be between 0 and 1.
The Eikonal loss $\cL_\text{eikonal}$ is also removed.
In~\figref{fig:ablation_representation} we compare reconstruction results with given GT poses, and can clearly see that using SDFs leads to more accurate geometry.

\tableablation
\figablationrepresentation

\section{Conclusions}\label{sec:conclusions}
We present NICER-SLAM, a novel dense RGB SLAM system that is end-to-end optimizable for both neural implicit map representations and camera poses.
We show that additional supervisions from easy-to-obtain monocular cues e.g. depths, normals, and optical flows can enable our system to reconstruct high-fidelity 3D dense maps and learn high-quality scene colors accurately and robustly in large indoor scenes.

\paragraph{Limitations}
Although we show benefits over SLAM methods using traditional scene representations in terms of mapping and novel view synthesis, our pipeline is not yet optimized to be real-time.
Also, no loop closure is performed under the current pipeline so the tracking performance should be further improvable.

{\small\paragraph{Acknowledgements}
 This project is partially supported by the SONY Research Award Program and a research grant by FIFA. The authors thank the Max Planck ETH Center for Learning Systems (CLS) for supporting Songyou Peng and the strategic research project ELLIIT for supporting Viktor Larsson. We thank Weicai Ye, Boyang Sun, Jianhao Zheng, and Heng Li for their helpful discussion.}

{\small
\bibliographystyle{ieee_fullname}
\bibliography{src/bib/reference}
}

\end{document}